\documentclass[letterpaper,9pt]{extarticle}
\usepackage{import}
\usepackage{local}
\usepackage[left=.7in,top=1in,bottom=1in,right=.7in]{geometry}

\usepackage{graphicx}
\usepackage{amsmath, bm}
\usepackage{siunitx}
\usepackage{multirow}
\usepackage{amsfonts}
\usepackage{booktabs}

\usepackage{tikz}
\newcommand{\dittotikz}{%
   \begin{tikzpicture}
        \draw [line width=0.12ex] (-0.2ex,0) -- +(0,0.8ex)
            (0.2ex,0) -- +(0,0.8ex);
        \draw [line width=0.08ex] (-0.6ex,0.4ex) -- +(-1.5em,0)
            (0.6ex,0.4ex) -- +(1.5em,0);
    \end{tikzpicture}%
}

\newcommand*\mean[1]{\bar{#1}}

\title{Hyperspectral unmixing for Raman spectroscopy via physics-constrained autoencoders}

\author{%
\textbf{Dimitar Georgiev,\textcolor{Accent}{\textsuperscript{1,2}} %
Álvaro Fernández-Galiana,\textcolor{Accent}{\textsuperscript{2,*}} %
Simon Vilms Pedersen,\textcolor{Accent}{\textsuperscript{2,3,*}}  %
Georgios Papadopoulos,\textcolor{Accent}{\textsuperscript{1,2}}  %
Ruoxiao Xie,\textcolor{Accent}{\textsuperscript{2}}  %
Molly M. Stevens,\textcolor{Accent}{\textsuperscript{2,4,\#}}  %
Mauricio Barahona\textcolor{Accent}{\textsuperscript{5,\#}} }\\[0.25cm]
\begin{small}
\textbf{\textcolor{Accent}{\textsuperscript{1}}}Department of Computing \& UKRI Centre for Doctoral Training in AI for Healthcare, Imperial College London, London, United Kingdom, SW7 2RH \\ 
\textbf{\textcolor{Accent}{\textsuperscript{2}}}Department of Materials, Department of Bioengineering \& Institute of Biomedical Engineering, Imperial College London, London, United Kingdom, SW7 2RH \\ 
\textbf{\textcolor{Accent}{\textsuperscript{3}}}Present address: SDU Biotechnology, Faculty of Engineering, University of Southern Denmark, Denmark, 5230 \\ 
\textbf{\textcolor{Accent}{\textsuperscript{4}}}Department of Physiology, Anatomy and Genetics, Department of Engineering Science, and Kavli Institute for Nanoscience Discovery, University of Oxford, Oxford, United Kingdom, OX1 3QU \\
\textbf{\textcolor{Accent}{\textsuperscript{5}}}Department of Mathematics, Imperial College London, London, United Kingdom, SW7 2AZ \\ 
\textbf{\textcolor{Accent}{\textsuperscript{*}}}Equal contribution. \\
\textbf{\textcolor{Accent}{\textsuperscript{\#}}}Correspondence: \textcolor{Accent}{molly.stevens@dpag.ox.ac.uk} \& \textcolor{Accent}{m.barahona@imperial.ac.uk}\\[0.25cm]
\textbf{\textcolor{Highlight}{Keywords:}} Raman spectroscopy, hyperspectral unmixing, chemometrics, autoencoders, machine learning\\[0.75cm]
\end{small}
}

\date{}
\begin{document}
\maketitle

\section{Abstract}

\begin{doublespacing}

\noindent
\textbf{\textcolor{Accent}{Raman spectroscopy is widely used across scientific domains to characterize the chemical composition of samples in a non-destructive, label-free manner. Many applications entail the unmixing of signals from mixtures of molecular species to identify the individual components present and their proportions, yet conventional methods for chemometrics often struggle with complex mixture scenarios encountered in practice. Here, we develop hyperspectral unmixing algorithms based on autoencoder neural networks, and we systematically validate them using both synthetic and experimental benchmark datasets created in-house. Our results demonstrate that unmixing autoencoders provide improved accuracy, robustness and efficiency compared to standard unmixing methods. We also showcase the applicability of autoencoders to complex biological settings by showing improved biochemical characterization of volumetric Raman imaging data from a monocytic cell.
}}

\section{Introduction}
Raman spectroscopy (RS) is a powerful optical modality that facilitates the identification, characterization and quantification of the molecular composition of chemical and biological specimens, offering in-depth insights into their structure and functionality~\cite{movasaghi2007raman, talari2015raman, butler2016using, mccreery2005raman, smith2019modern}. 
RS interrogates the vibrational modes of molecules through the analysis of inelastic scattering of monochromatic light from matter, thereby enabling the non-destructive, label-free fingerprinting of chemical species~\cite{koningstein2012introduction, szymanski2012raman, colthup2012introduction, jones2019raman, bocklitz2016raman}. 
As a result, RS has become an important analytical tool in a myriad of scientific domains, from chemistry~\cite{schlucker2014surface, dodo2022raman}, biology~\cite{pezzotti2021raman, smith2016raman, shipp2017raman, cialla2017recent}, and medicine~\cite{kong2015raman, ember2017raman, pence2016clinical, balan2019vibrational, auner2018applications, mahadevan1996raman, tanwar2021advancing}, to materials science~\cite{fernandez2023fundamentals, kumar2012raman}, pharmacology~\cite{wang2018research, paudel2015raman, VANKEIRSBILCK2002869}, environmental science~\cite{halvorson2010surface, ong2020surface, terry2022applications}, food quality control~\cite{li2014raman, pang2016review}, and even forensics~\cite{chalmers2012infrared, khandasammy2018bloodstains, izake2010forensic}.

Despite the wealth of information RS affords, the analysis and interpretation of experimental RS data remains a major challenge~\cite{ryabchykov2018analyzing, guo2021chemometric, gautam2015review}. Many important applications entail the analysis of complex mixtures of molecular species coexisting and interacting at micro- and nanoscales. Such complexity can hinder the qualitative and quantitative investigation of RS measurements, especially when dealing with the biomolecular diversity of biological samples~\cite{byrne2016spectral, gautam2015review}.

Hyperspectral unmixing (also known as (hyper)spectral deconvolution or multivariate curve resolution) aims to resolve such mixed signals~\cite{li2017spectral, olmos2017relevant} by identifying the individual components present (\textit{endmember identification}) and/or quantifying their proportions (\textit{abundance estimation}) (see Fig.~\ref{fig:diagram}a). Popular approaches include N-FINDR~\cite{nfindr} and Vertex Component Analysis (VCA)~\cite{vca} for endmember identification, and Non-negative Least Squares (NNLS)~\cite{nnls} and Fully Constrained Least Squares (FCLS)~\cite{fcls} for abundance estimation~\cite{li2017spectral, hedegaard2011spectral}. However, such techniques, which originated in the field of remote sensing~\cite{keshava2002spectral, harris2006spectral}, have limitations for the unmixing of RS data. Specifically, these methods are typically restricted to linear mixing; lack robustness to data artifacts abundant in RS data (e.g., dark noise, baseline variations, cosmic spikes);  rely on additional assumptions (e.g., endmembers present as `pure pixels' in the data) or require additional information (e.g., number of endmembers, underlying mixture model, endmember library); and are computationally demanding for large datasets (e.g., imaging and volumetric Raman raster scans). 

Autoencoder (AE) neural networks have recently emerged as a framework to enhance the precision of hyperspectral unmixing in remote sensing, spurred by the availability of standardized benchmark datasets for model evaluation (e.g., \textit{Urban}, \textit{Samson}, \textit{AVARIS Cuprite})~\cite{palsson2022blind,zhang2020recent, wang2022self, bhatt2020deep,chen2022integration}. Yet, despite initial explorations~\cite{burzynski2021deep, boildieu2023multivariate}, the utility of unmixing AEs for Raman spectroscopy data remains largely unexplored. Here, we develop a range of AEs for RS hyperspectral unmixing, which we systematically validate against conventional unmixing methods using synthetic and experimental Raman data.

\section{Results}
\subsection*{Raman unmixing with autoencoder neural networks}
Autoencoders are a family of (deep) neural network models consisting of two sub-networks (\emph{encoder} and \emph{decoder}) connected sequentially ~\cite{goodfellow2016deep}. The encoder $\mathcal{E}: \mathbb{R}^b \to\mathbb{R}^m$, where $m \ll b$, transforms input data $\mathbf{x}$ to a lower-dimensional latent space representation $\mathbf{z}=\mathcal{E}(\mathbf{x})$, which the decoder $\mathcal{D}: \mathbb{R}^m \to\mathbb{R}^b$ uses to produce reconstructions $\widehat{\mathbf{x}}= \mathcal{D}(\mathbf{z})$ of the original input. AE models are typically trained in a \emph{self-supervised} manner by minimizing a loss function $\mathcal{L}(\mathbf{x}, \widehat{\mathbf{x}})$ that measures the discrepancy between the input $\mathbf{x}$ and the reconstruction $\widehat{\mathbf{x}}$ (e.g., the mean squared error (MSE)). Therefore, no ground truth information is provided during training as opposed to previously reported machine learning approaches for Raman hyperspectral analysis and characterization based on supervised learning~\cite{manifold2021versatile, zhang2020high, wei2022pixel}. As the training of the model proceeds, the encoder progressively learns a latent representation that captures the most salient features of the input data, whereas the decoder learns how to recover the data back from the latent representation. 

\begin{figure}[!t]
    \centering
    \includegraphics[width=17.8cm]{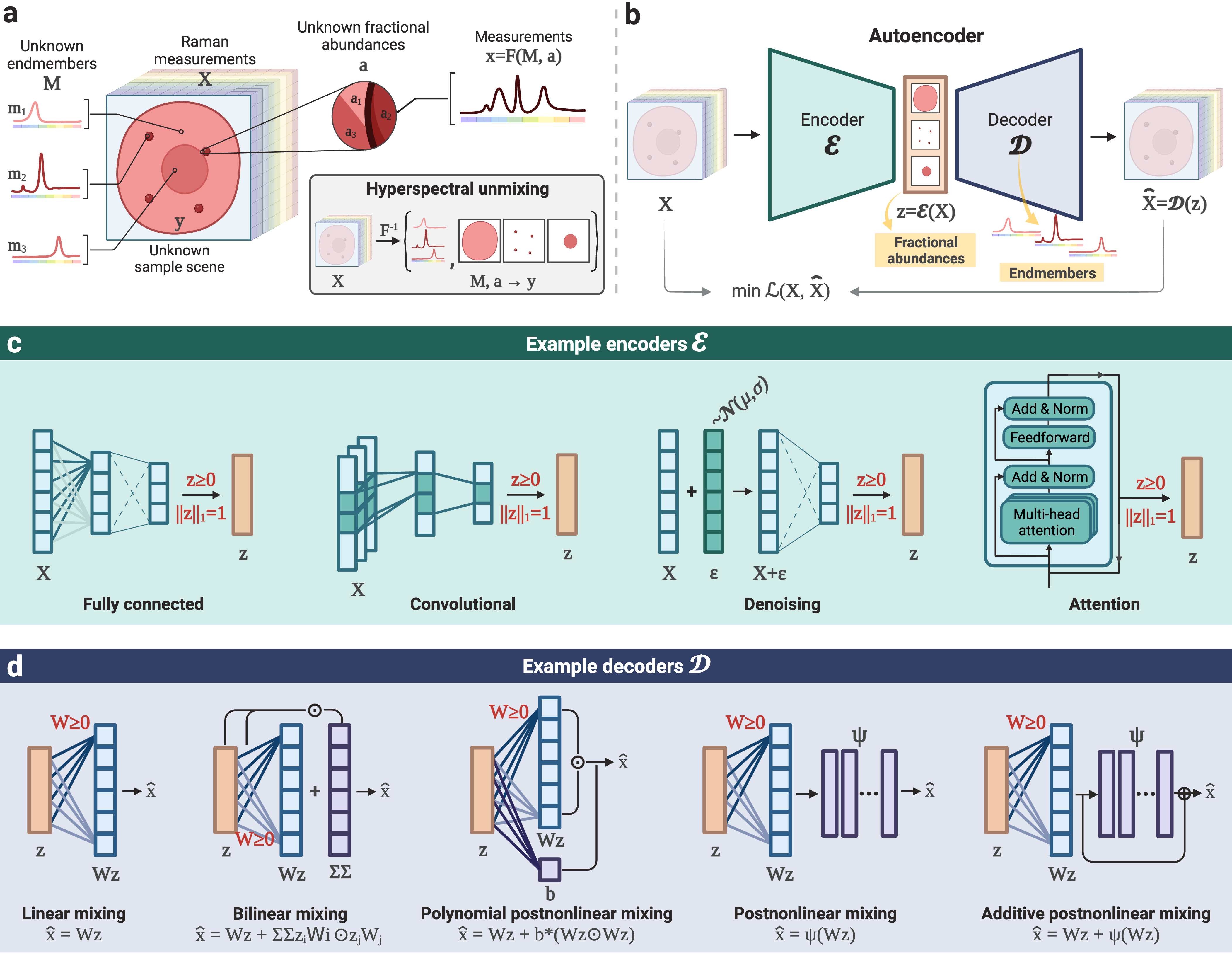} 
    \caption{\textbf{Hyperspectral unmixing for Raman spectroscopy using autoencoder neural networks.} \textbf{a, } Diagram of the task of hyperspectral unmixing. 
    \textbf{b, } Hyperspectral unmixing as a self-supervised autoencoder learning problem: the decoder learns to derive endmembers and the encoder learns the corresponding fractional abundances. \textbf{c, } Encoders can accommodate different concepts from representation learning, such as convolutional layers and attention, to improve feature extraction and provide more accurate and robust unmixing. \textbf{d, } Decoders can be structured to model different linear and non-linear mixing models. Labels in red in \textbf{c} and \textbf{d} indicate physics-inspired constraints.}
    \label{fig:diagram}
\end{figure}

This dual functionality can be harnessed to design autoencoders for hyperspectral unmixing: the latent representations $\mathbf{z}=\mathcal{E}(\mathbf{x})$ can be interpreted as fractional abundances (with respect to the input spectrum $\mathbf{x}$), and the decoder $\mathcal{D}(\cdot)$ acts as a mixing function on these representations by encoding endmember signatures and other interactions. Hence, AE models learn to perform an `unmixing' where the decoder identifies endmember signatures and the encoder quantifies the fractional abundances of these learned endmembers in the input spectrum (Fig.~\ref{fig:diagram}b). To guide the learning, we incorporate physical constraints into the AE architecture to reflect the nature of hyperspectral unmixing, e.g., non-negativity of endmembers and fractional abundances, and sum-to-one abundances (Fig.~\ref{fig:diagram} and \textit{Materials and Methods}). 

This setup provides a more adaptable and versatile framework for unmixing, which can address many of the limitations of conventional techniques (as summarized in Table~\ref{tab:advantages}). On the one hand, the learning of physical and biochemical features in the encoder can be enhanced by adopting strategies from representation learning, such as convolutional layers to capture spectral and/or spatial correlations among neighboring bands and/or pixels~\cite{zhang2018hyperspectral, palsson2020convolutional, elkholy2020hyperspectral}, or attention mechanisms to model long-range dependencies~\cite{ghosh2022hyperspectral} (Fig.~\ref{fig:diagram}c). In addition, sparsity, part-based learning and denoising objectives can be enforced during training to enhance explainability and robustness~\cite{ozkan2018endnet, qu2018udas, su2018stacked, su2017nonnegative, qu2017spectral}.

On the other hand, the design of the decoder allows for flexible modeling of input data, specifically to account for various mixture models, e.g., linear, bilinear and post-nonlinear (Fig.~\ref{fig:diagram}d)~\cite{chen2022integration, shahid2021unsupervised, zhao2021hyperspectral}. This is akin to introducing an inductive prior with respect to the mixture model directly via the AE architecture. Furthermore, the decoder can be pre-initialized with a set of endmembers (e.g., an endmember library or signatures derived using methods such as VCA), or readily adapted to non-blind unmixing by fixing certain parameters to predefined endmember signatures.

\subsubsection*{Unmixing autoencoders}
To assess the effectiveness of AEs as a framework for RS unmixing, we develop and evaluate a collection of AE models, each defined by a specific encoder and decoder. We consider four types of encoders encompassing a variety of architectures, from traditional dense layers to contemporary convolutional and attention mechanisms: 1) an encoder consisting of fully connected layers (\emph{Dense}); 2) an encoder with a 1D convolutional feature extractor block, followed by a fully connected part (\emph{Convolutional}); 3) a transformer-based encoder that uses multi-head attention (\emph{Transformer})~\cite{vaswani2017attention}; and 4) a transformer-based encoder with a 1D convolutional feature extractor (\emph{Convolutional Transformer}). The two types of decoders we investigate are: 1) a decoder designed for linear unmixing (Eq.~\ref{eq:ae_linear_mixing}); and 2) a decoder designed for bilinear unmixing (Eq.~\ref{eq:fan_autoencoder}). The autoencoders are trained in a self-supervised fashion by minimizing a loss based on the spectral angle divergence (SAD)~\cite{sad} that measures the cosine similarity between input and reconstructed spectra.

\subsubsection*{Baseline methods for comparison}
We compare AE performance to conventional unmixing approaches: N-FINDR and VCA as endmember extraction algorithms followed by NNLS or FCLS to derive fractional abundances. This is performed using the \emph{RamanSPy} package~\cite{georgiev2023ramanspy} in Python. We also compared to Principal Component Analysis (PCA), which, despite not being designed for unmixing, is commonly used in applications. We omit the PCA results in the main text as they exhibit substantially lower performance (see SI).

\subsection*{Benchmarking unmixing autoencoders on synthetic Raman mixtures}
We first benchmark the performance of our AE architectures on synthetic datasets created in-house.

\begin{figure*}
    \centering
    \includegraphics[width=17.8cm]{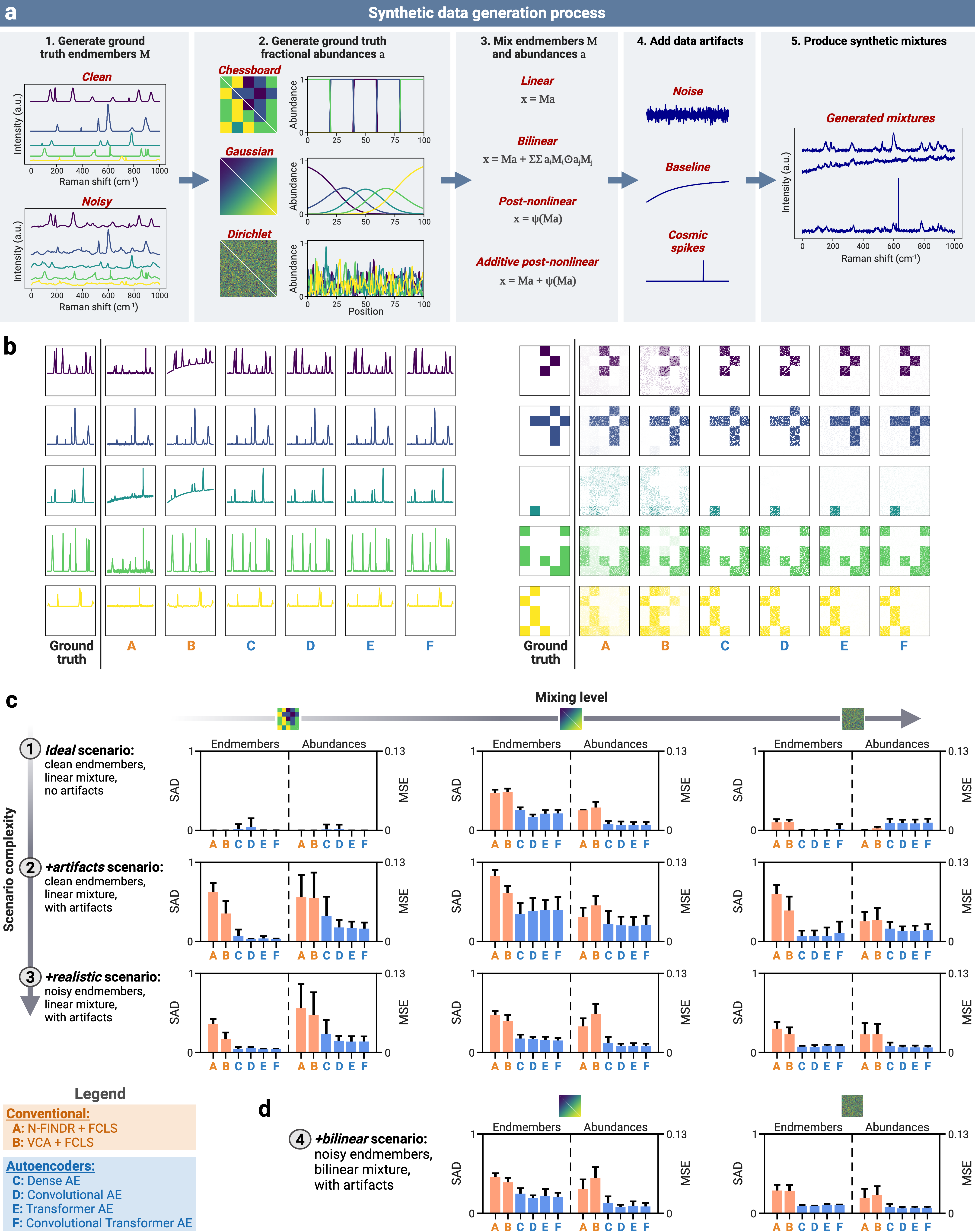}
    \caption{\textbf{Benchmarking autoencoders on synthetic Raman mixtures.} 
    \textbf{a,} Schematic of our synthetic data generation workflow. 
    \textbf{b,} Representative results for the six algorithms (two conventional and four AEs) on an example synthetic dataset (\emph{Chessboard}\emph{+artifacts} scenario): endmembers (\textit{left}), and fractional abundances (\textit{right}). 
    \textbf{c-d,} Summary of unmixing performance on synthetic datasets of variable mixing level and complexity: linear mixtures (\textbf{c}), bilinear mixtures (\textbf{d}).
    Confidence intervals are given by one standard deviation around the sample mean ($n=25$ samples: $5$ datasets with $5$ model repetitions each).
    }
    \label{fig:synthetic_results}
\end{figure*}

\subsubsection*{Synthetic data generation}
We developed a custom data generator that produces synthetic Raman mixtures with different characteristics (e.g., number and type of endmembers, abundance profiles, mixture model, data artifacts) with full knowledge of the `ground truth' endmembers and fractional abundances (Fig.~\ref{fig:synthetic_results}a). This allows us to compare the performance of unmixing approaches (see Fig.~\ref{fig:synthetic_results}b for unmixing of an example synthetic dataset).

Using our data generator, we produce $11$ types of synthetic datasets of variable complexity, based on four mixture scenarios over three fractional abundance scenes. In order of complexity, the four mixture scenarios are: 1) a linear mixture with \emph{clean} endmembers and no data artifacts (\emph{ideal}); 2) a linear mixture with \emph{clean} endmembers, but contaminated with artifacts representing dark noise, baseline variations and cosmic spikes  (\emph{+artifacts}); 3) a linear mixture with \emph{noisy} endmembers (i.e., containing additional smaller noise peaks) and artifacts (\emph{+realistic}); and 4) a bilinear mixture based on the Fan model~\cite{fan2009comparative} with \emph{noisy} endmembers and artifacts (\emph{+bilinear}). For each of the four mixture scenarios, we generate three dataset variants (two for the \emph{+bilinear} scenario since no bilinear interactions are present in our \emph{Chessboard} scene) based on custom $100\times100$ fractional abundance scenes. This produces $10$k spectra per dataset, organized into two-dimensional scenes for visualization purposes. In increasing level of mixing, we have: 1) a scene comprising well-separated patches, each containing a single species (\emph{Chessboard} scene); 2) a semi-mixed scene given by a Gaussian mixture of species (\emph{Gaussian} scene); and 3) a highly-mixed scene where each pixel represents a random sample of species drawn from a Dirichlet distribution (\emph{Dirichlet} scene). Therefore, our synthetic datasets cover increasingly complex scenarios, from the \emph{ideal} \emph{Chessboard} dataset, which is trivial for conventional methods, to noisier, more complex mixtures containing different types of artifacts.

\subsubsection*{Benchmark results on linear mixtures}
We first discuss our results on the nine dataset variants created through the linear mixture scenarios (1-3). Such data complies with the linear mixing assumption of conventional methods and, for consistency, we equip the AE models with a decoder for linear unmixing. Fig.~\ref{fig:synthetic_results}c summarises the performance of the six models (two conventional and four AEs) across the nine dataset variants, with experiments performed over $5$ distinct datasets and $5$ model initializations for each variant (refer to Table~\ref{tab:synthetic_metrics} for calculated performance metrics). We measure the discrepancy between ground truth and derived endmembers (using SAD), and the discrepancy between ground truth and derived fractional abundances (using MSE). We find that the AE models outperform the two conventional methods, providing more accurate endmembers and fractional abundances across virtually all scenarios and abundance scenes. The AEs recover the performance of the conventional methods on the simple \emph{ideal} \emph{Chessboard} datasets, and the improvement in AE performance becomes increasingly prominent for mixture scenarios with higher levels of noise and data artifacts.

\subsubsection*{Non-linear unmixing with autoencoders}
We then proceed to our benchmark analysis on synthetic data generated using a non-linear mixture model (i.e., \emph{+bilinear} scenario). The results are displayed in Fig.~\ref{fig:synthetic_results}d, where we equip our AEs with a decoder specific to the bilinear mixture model by merely adapting the decoder architecture. Again, we observe that all four AE models provide a substantial improvement in unmixing accuracy compared to methods like N-FINDR+FCLS and VCA+FCLS for both endmember and abundance estimation.

\begin{figure}[!b]
    \centering
    \includegraphics[width=8.7cm]{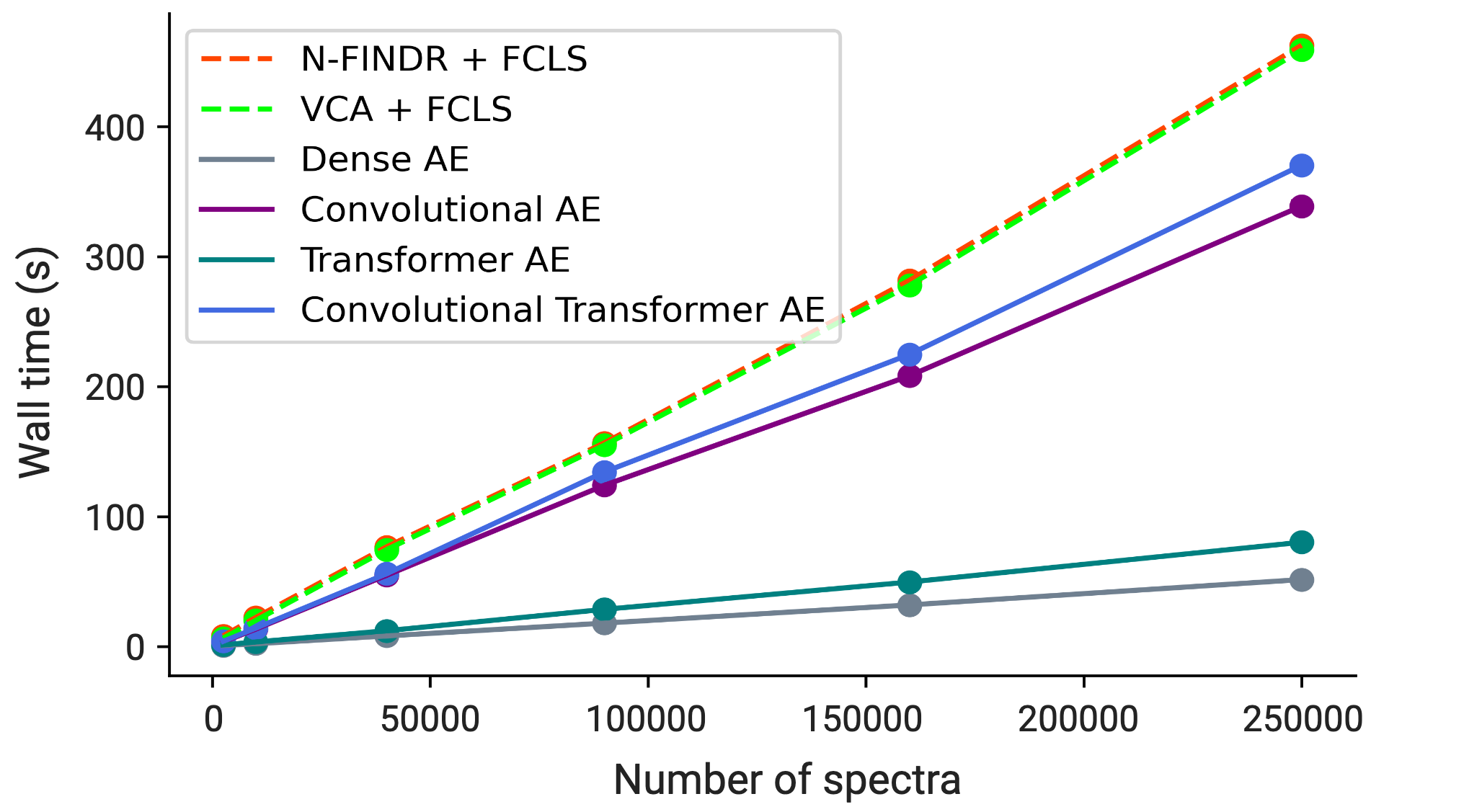}
    \caption{\textbf{Computational efficiency of autoencoders and conventional methods on synthetic datasets with an increasing number of spectra.} Each dot represents the average across 3 evaluations (confidence intervals based on one standard deviation are small and not visible to the eye). AE models are equipped with decoders for linear unmixing. Data generated under \emph{Chessboard +artifacts}.
    }
    \label{fig:comlexity}
\end{figure}

\subsubsection*{Computational efficiency}
The computational complexity and scalability of unmixing methods can become a significant bottleneck in real-world applications, particularly for imaging and volumetric Raman scans, which can contain hundreds of thousands of spectra. To examine this issue, we profile the computational cost of our four AE methods (linear decoders) and the two conventional methods on synthetic datasets of increasing size up to $250000$ spectra. To be fair to conventional algorithms, we include the full training time for autoencoders and we use standard CPU computation to avoid any advantage from GPU acceleration. Fig.~\ref{fig:comlexity} shows that all AE models are faster than N-FINDR+FCLS and VCA+FCLS, which are already among the most computationally lightweight conventional unmixing techniques~\cite{bioucas2012hyperspectral}. Hence, AEs provide efficient unmixing, even without utilizing GPU acceleration and parallel processing, which can further enhance their performance.

\begin{figure*}[!t]
    \centering
    \includegraphics[width=17.8cm]{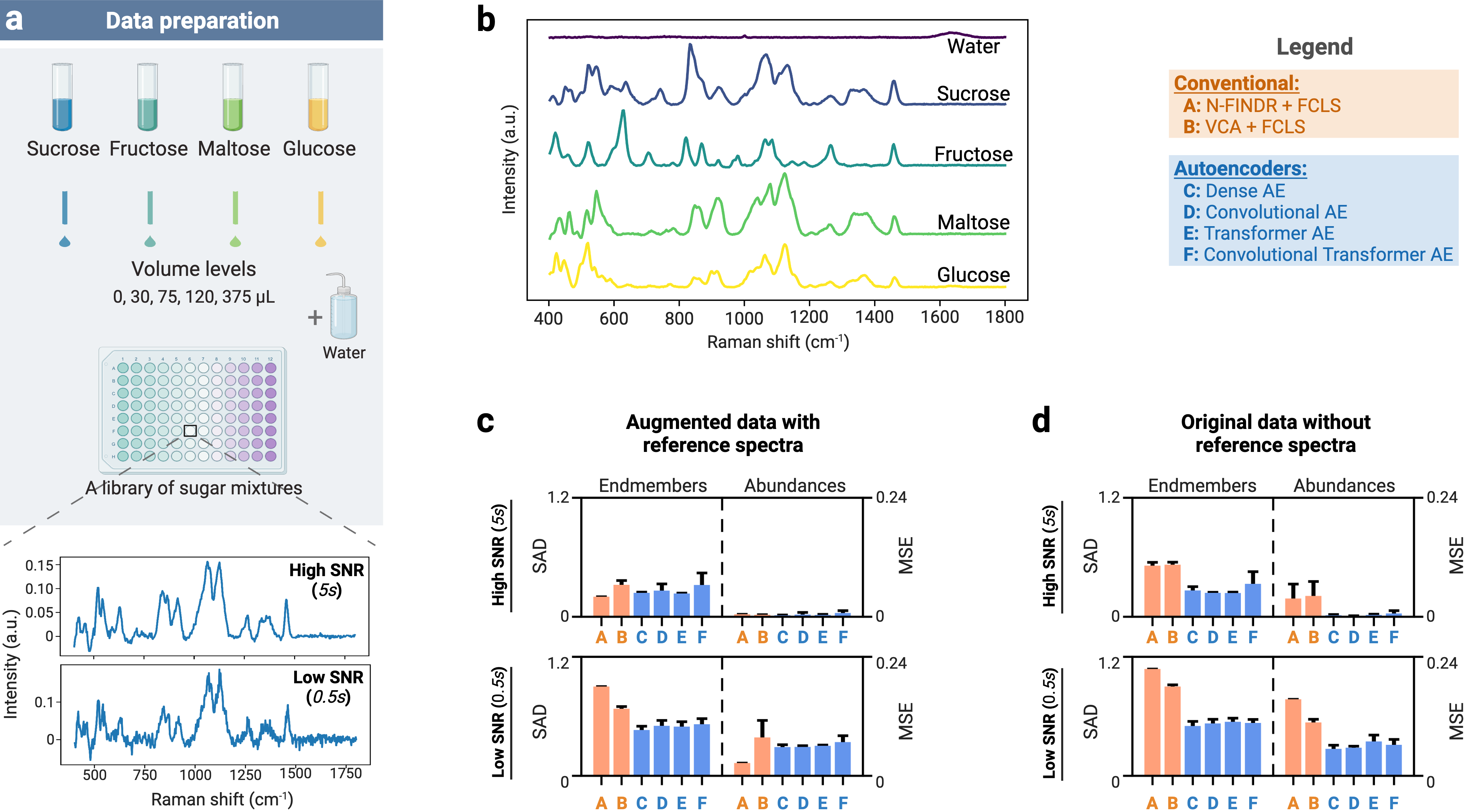}
    \caption{\textbf{Experimental validation on Raman spectroscopy data from sugar solutions.} 
    \textbf{a,} Schematic diagram of sugar mixture preparation. Two sets of data are acquired---high and low signal-to-noise ratio (SNR) data, by using integration times of \qty{5}{\s} and \qty{0.5}{\s}, respectively. 
    \textbf{b,} Endmember signatures estimated from 
    reference spectra (high SNR) additionally collected from pure solutions.
    \textbf{c-d,} Summary of unmixing performance for: \textbf{c,} idealized scenario with augmented data including reference spectra; and \textbf{d,} original data without augmentation. Confidence intervals are given as one standard deviation around the sample mean ($n=5$).
    }
    \label{fig:sugar_data}
\end{figure*}

\subsection*{Validation of unmixing autoencoders on experimental Raman data from sugar mixtures}
To validate the unmixing performance of AEs on real experimental data, we next performed benchmark analyses on data from a library of $240$ sugar mixtures prepared in-house with four types of sugar (glucose, sucrose, fructose, maltose) at different concentrations (see Fig.~\ref{fig:sugar_data}a). To evaluate different signal-to-noise (SNR) conditions, we acquired high SNR ($1920$ spectra) and low SNR ($7680$ spectra) measurements using a custom Raman microspectroscopy platform at integration times of \qty{5}{\s} and \qty{0.5}{\s}, respectively. We used these experimental datasets with ground truth to systematically evaluate unmixing algorithms under typical experimental artifacts, such as baseline shifts, environmental noise, and cosmic spikes.  
 
We perform unmixing on these data to identify the content of each mixture, i.e., types of sugar and their concentrations. The ground truth is defined by the experimental concentrations and the endmember signatures we obtain from reference spectra measured from $5$ additional pure solutions (Fig.~\ref{fig:sugar_data}b). As with the synthetic data above, we benchmark the performance of our four AE models (linear decoders) against N-FINDR+FCLS and VCA+FCLS. 

First, we consider an idealized scenario, purposefully devised to favor conventional methods, whereby endmembers are present in the data. To do this, we augment our data with the additional reference spectra we measured. When such `pure pixels' are available, we observe that conventional methods (NFINDR+FCLS, VCA+FCLS) perform comparably to AEs on clean, high SNR data (Fig.~\ref{fig:sugar_data}c). Yet, AEs already provide improved performance in low SNR regimes.

\begin{figure*}[!h]
    \centering
    \includegraphics[width=17.8cm]{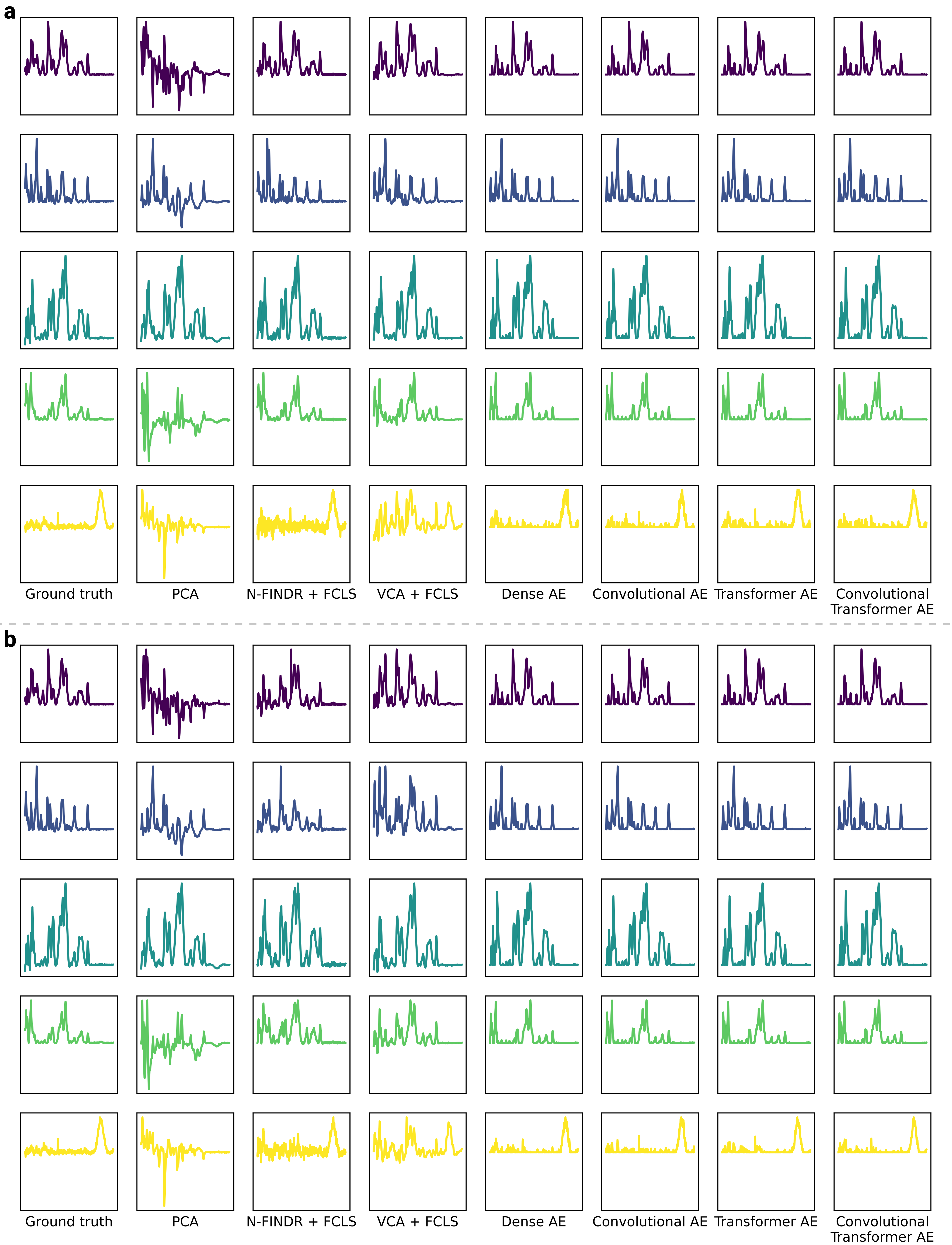} 
    \caption{\textbf{Endmember estimates on high SNR sugar data.} Qualitative comparison of derived endmembers on the high SNR sugar datasets - with (\textbf{a}), and without reference spectra (\textbf{b}). Endmembers are scaled such that their maximum intensity is equal to $1$ for visualization purposes. \textit{x}-axes represent the Raman shift region \numrange[range-phrase=--]{400}{1800}\,\unit{\per\centi\metre}, and \textit{y}-axes, which are shared for each row, represents normalized intensity (a.u.).
    }
    \label{fig:hsnr_representative}
\end{figure*}

\begin{figure*}
    \centering
    \includegraphics[width=17.8cm]{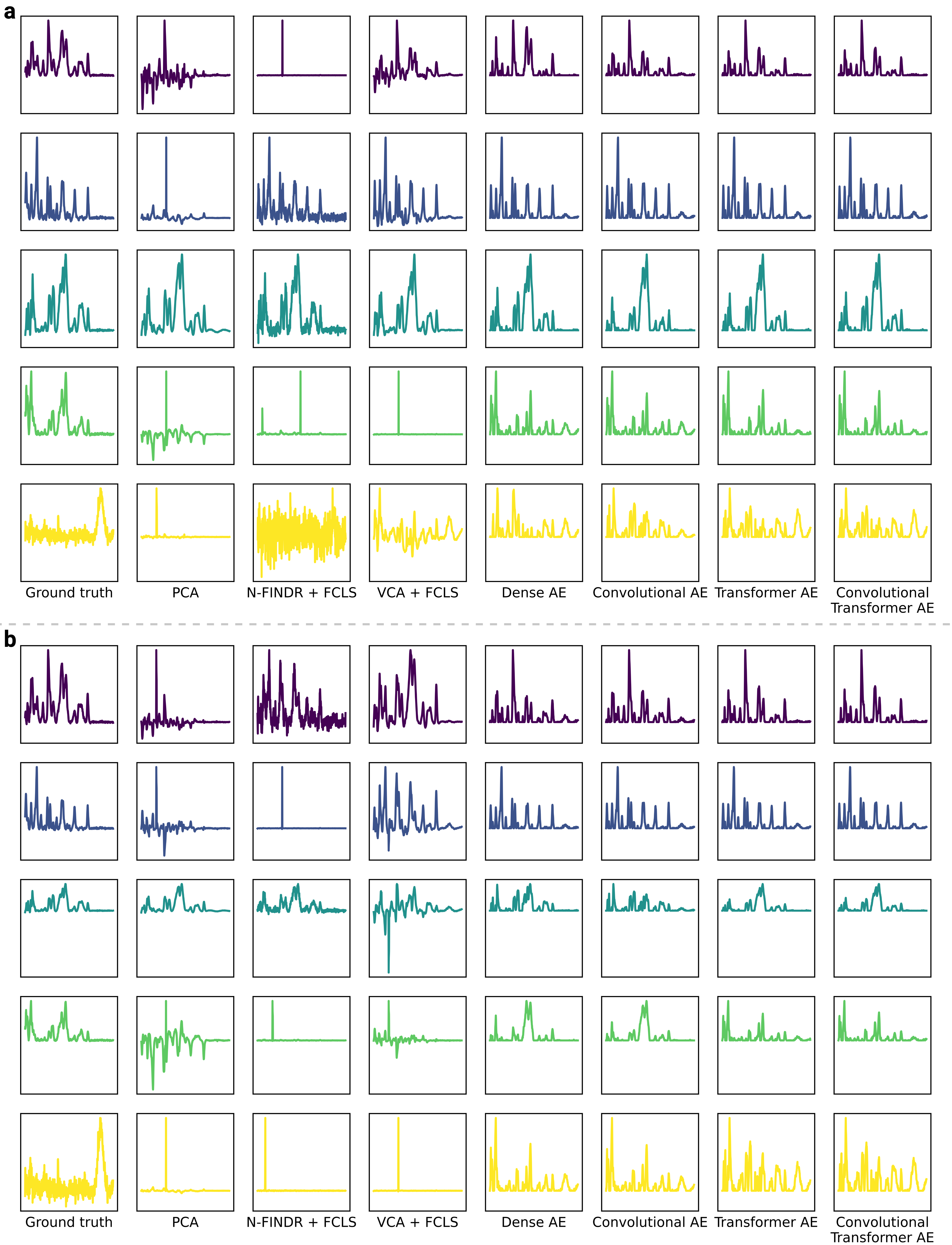} 
    \caption{\textbf{Endmember estimates on low SNR sugar data.} Qualitative comparison of derived endmembers on the low SNR sugar datasets - with (\textbf{a}), and without reference spectra (\textbf{b}). Endmembers are scaled such that their maximum intensity is equal to $1$ for visualization purposes. \textit{x}-axes represent the Raman shift region \numrange[range-phrase=--]{400}{1800}\,\unit{\per\centi\metre}, and \textit{y}-axes, which are shared for each row, represents normalized intensity (a.u.).
    }
    \label{fig:lsnr_representative}
\end{figure*}

In many experimental applications, however, the underlying endmembers are not present in the data and cannot be separately obtained (e.g., target-agnostic applications, or unknown species). To consider such cases, we analyzed our original datasets without augmentation. Our results in Fig.~\ref{fig:sugar_data}d demonstrate that, in such situations, AEs substantially outperform conventional methods in both low and high SNR settings (see Figs.~\ref{fig:hsnr_representative}-\ref{fig:lsnr_representative} for additional data).

\begin{figure}[!t]
    \centering
    \includegraphics[width=17.8cm]{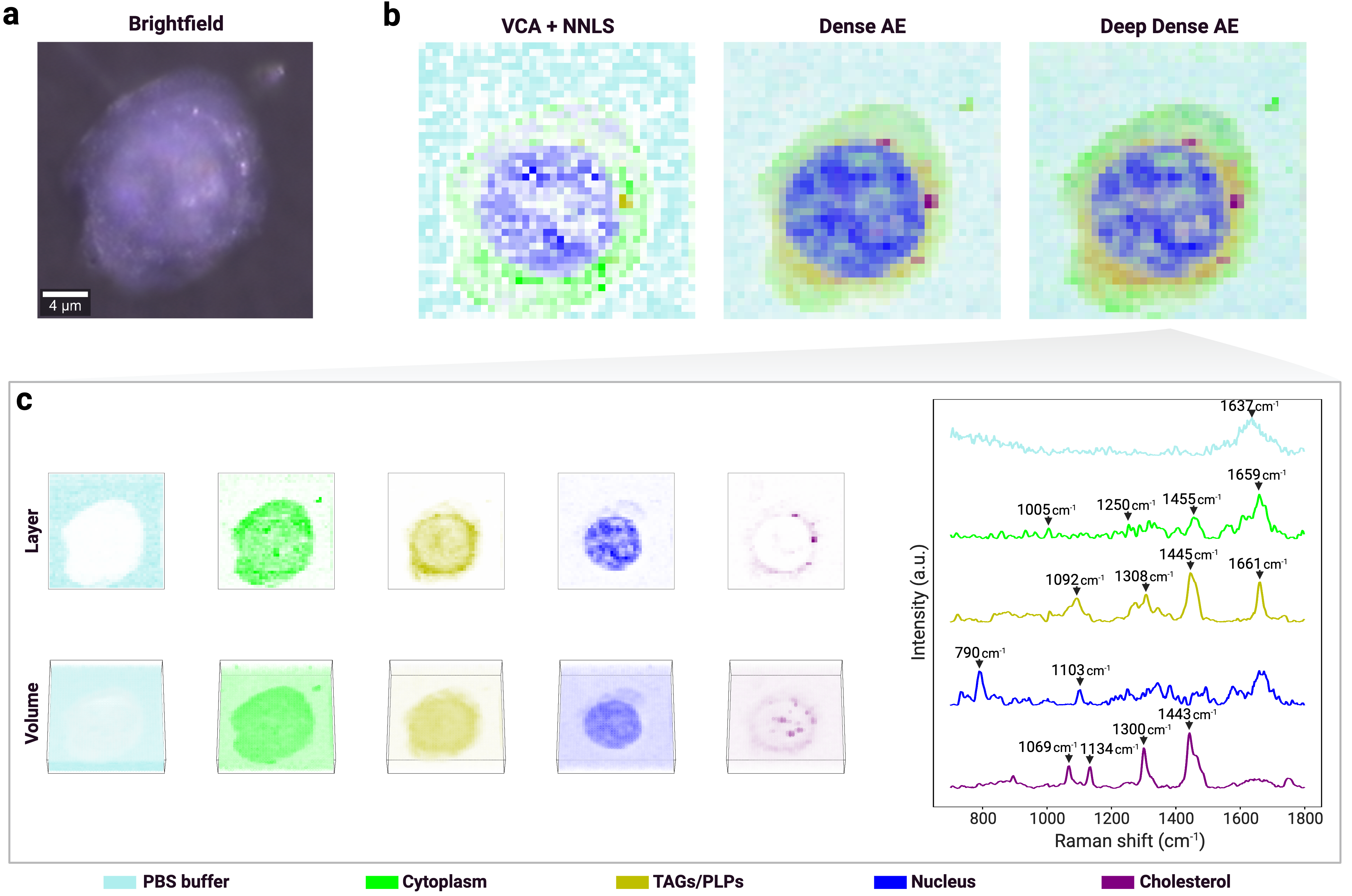} 
    \caption{\textbf{Analysis of volumetric Raman imaging of a THP-1 cell with unmixing autoencoders.} \textbf{a,} A brightfield image of the studied THP-1 cell. \textbf{b,} A cross-section reconstruction of the cell (layer $z=7$) obtained by overlaying the fractional abundances derived by: VCA+NNLS, our \emph{Dense AE}, and a deeper \emph{Dense AE}. \textbf{c,} Results obtained with our deeper \emph{Dense AE} model, displaying the spatial distribution of the individual fractional abundances and the associated endmember signatures. Fractional abundance maps normalized for consistent visualization. Data from Kallepitis~\textit{et~al.}~\cite{kallepitis2017quantitative}.}
    \label{fig:real_application}
\end{figure}

\subsection*{Application of unmixing autoencoders to biological data: volumetric Raman imaging of a THP-1 cell}

As an application to biological research, we use unmixing autoencoders to analyze a low-SNR volumetric RS raster scan of a human leukemia monocytic (THP-1) cell (Fig.~\ref{fig:real_application}a)~\cite{kallepitis2017quantitative}. Using Raman chemometrics, the composition of the cell can be probed to study its morphology in a non-destructive, label-free manner. 

After loading and preprocessing the data using \emph{RamanSPy}~\cite{georgiev2023ramanspy}, we conduct hyperspectral unmixing with: 1) VCA+NNLS - as in the original paper; 2) \emph{Dense AE} - our simplest and computationally efficient AE model; and 3) \emph{Deep Dense AE} - an extension of \emph{Dense AE} with a deeper encoder with five layers. We derive $20$ endmembers, which we characterize via peak assignment to identify biochemical species present in the scanned cell, such as deoxyribonucleic acid (DNA), proteins, triglycerides (TAGs), phospholipids (PLPs) and cholesterol esters (see \textit{Methods and Materials}).

\begin{figure*}
    \centering
    \includegraphics[width=17.8cm]{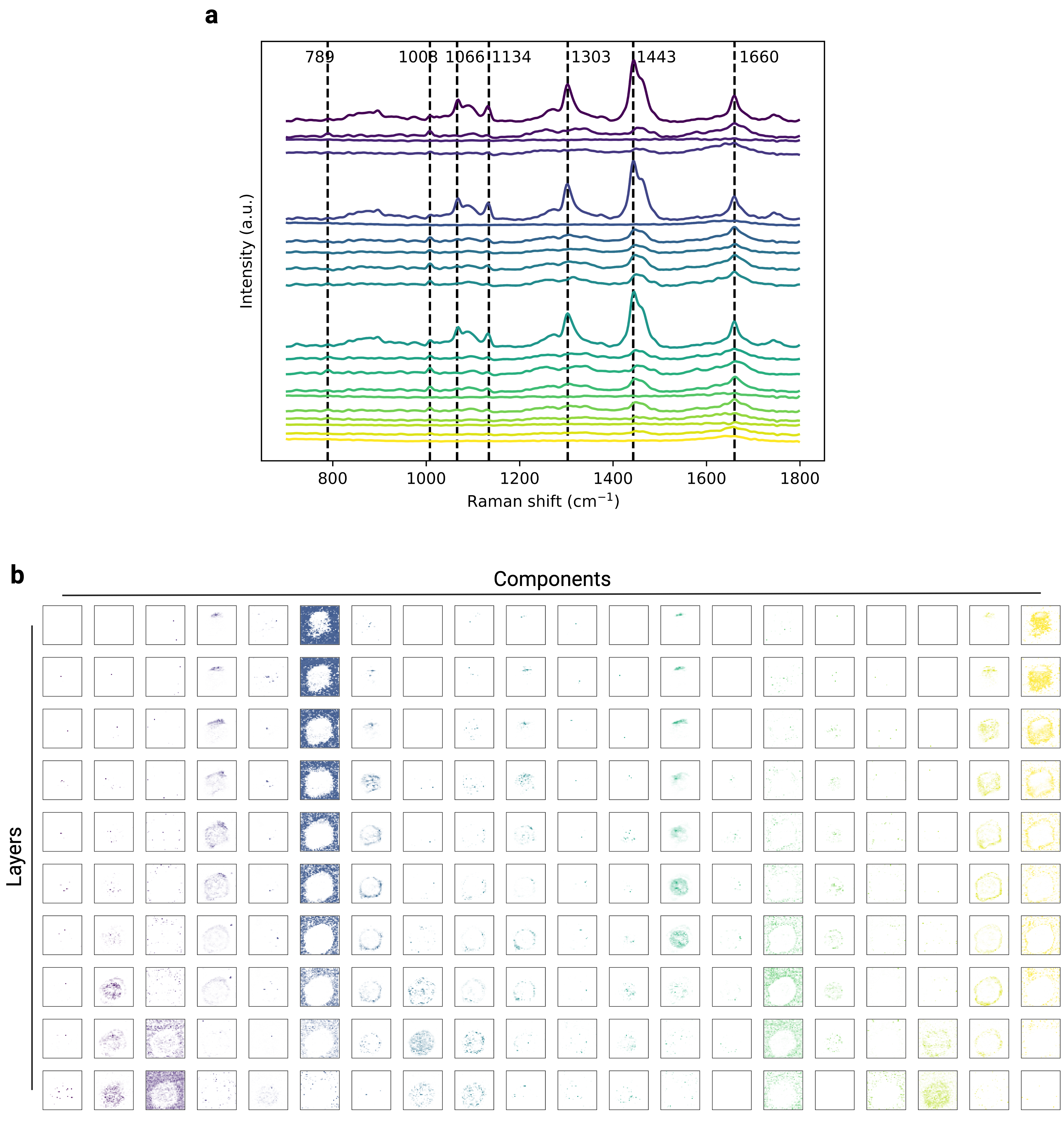}
    \caption{\textbf{Full unmixing results obtained with VCA + NNLS on the THP-1 cell data.} \textbf{a,} Derived endmembers. \textbf{b,} Derived fractional abundances.
    }
    \label{fig:thp1_vca_full}
\end{figure*}

\begin{figure*}
    \centering
    \includegraphics[width=17.8cm]{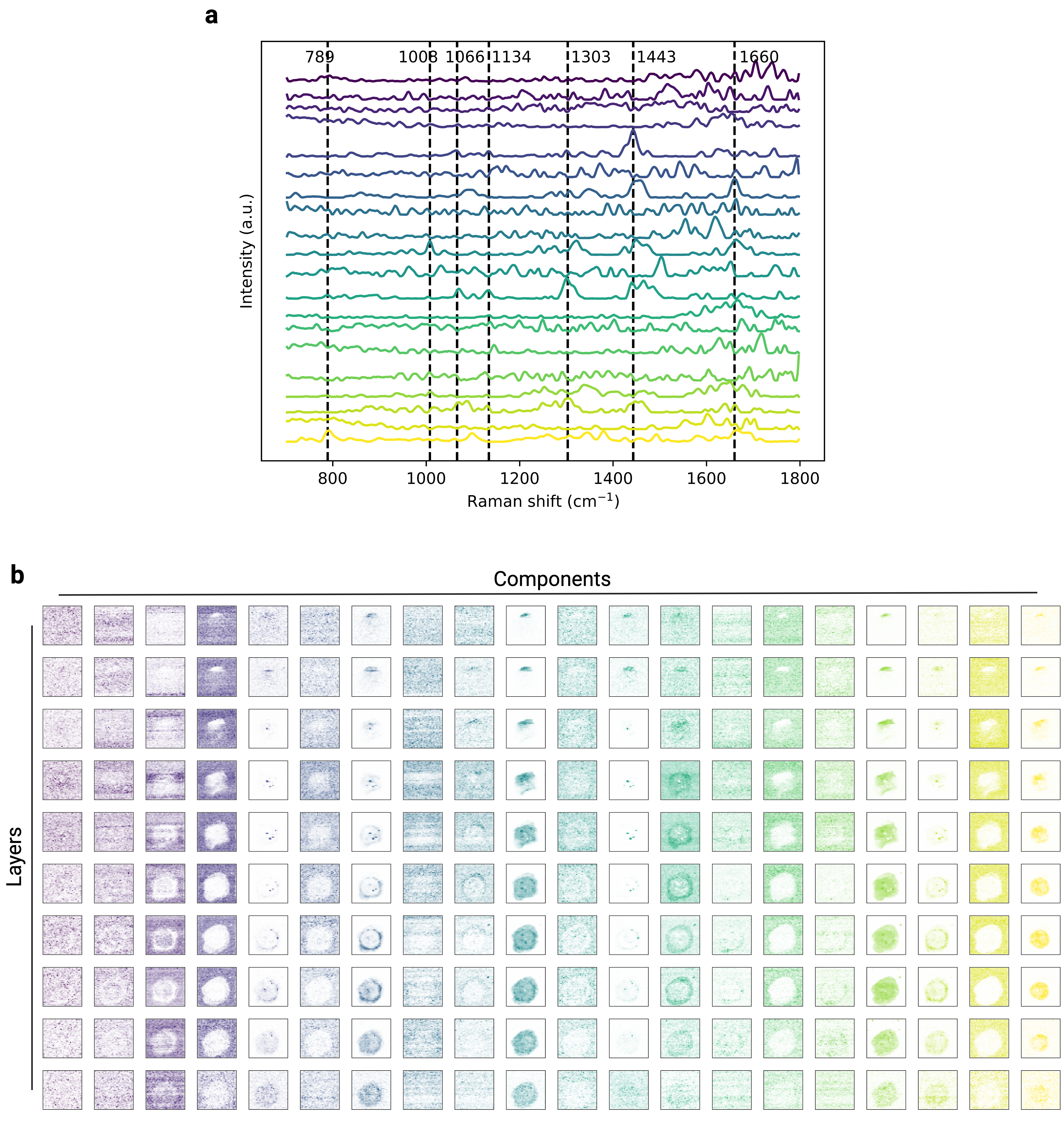}
    \caption{\textbf{Full unmixing results obtained with our \emph{Dense AE} model on the THP-1 cell data.} \textbf{a,} Derived endmembers. \textbf{b,} Derived fractional abundances.
    }
    \label{fig:thp1_dense_ae_full}
\end{figure*}

\begin{figure*}
    \centering
    \includegraphics[width=17.8cm]{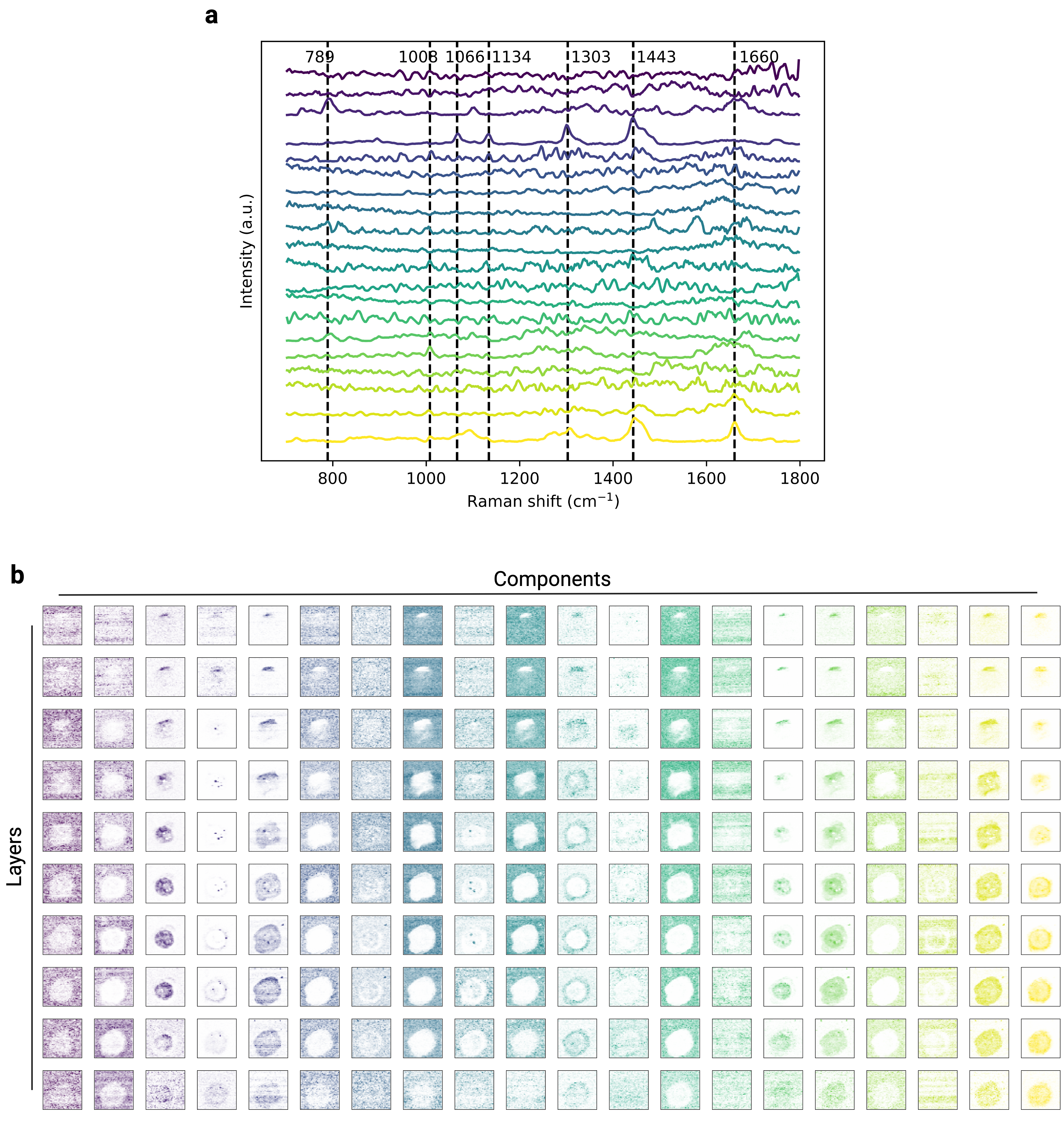}
    \caption{\textbf{Full unmixing results obtained with our \emph{Deep Dense AE} model on the preprocessed THP-1 cell data.} \textbf{a,} Derived endmembers. \textbf{b,} Derived fractional abundances.
    }
    \label{fig:thp1_deep_dense_ae_full}
\end{figure*}

Fig.~\ref{fig:real_application}b shows the reconstructions of the cell created by overlaying selected fractional abundances derived by each method, revealing the spatial organization of key cellular organelles, including the nucleus, cytoplasm, lipid bodies and membranes. Although direct comparisons are challenging due to the lack of ground truth, the unmixing results of our AE models are aligned with the original findings~\cite{kallepitis2017quantitative}, albeit with more distinct endmember signatures and well-defined abundance features (see Figs.~\ref{fig:thp1_vca_full}-\ref{fig:thp1_deep_dense_ae_full} for full results). The \emph{Deep Dense AE} model provides cleaner endmember signatures that enable more precise spectral and compositional information (Fig.~\ref{fig:real_application}c-d). Notably, unlike the original VCA+NNLS approach, our AEs detect cholesterol, an important functional and structural component in cells, where it plays a key role in membrane fluidity and stability, signaling pathways, and immune response~\cite{kritharides1998cholesterol, tall2015cholesterol, saha2017cellular}. 

\section*{Conclusion}
In this work, we have presented an autoencoder-based methodology for hyperspectral unmixing in Raman spectroscopy, which we validated on a wide array of synthetic and real experimental datasets. Our results demonstrate that autoencoders are adept at handling diverse mixture scenarios and exhibit robustness against data artifacts, offering an effective, versatile and efficient framework for RS unmixing.

The potential of autoencoders for RS unmixing opens several avenues for future research. One direction is the investigation of AE architectures with more complex decoders~\cite{chen2022integration} and/or encoders (e.g., stacked and denoising AE architectures~\cite{qu2017spectral, su2018stacked}), as well as the use of training objectives that better capture spectral reconstruction. Another promising area is the use of AEs as a pretraining procedure in downstream tasks, potentially combined with other AI-based approaches (e.g., deep learning models for denoising~\cite{horgan2021high}).

Finally, while our focus here is on RS, we wish to underscore the applicability of our work to other spectroscopic modalities, including infrared spectroscopy.

\section{Materials and Methods}
\subsection*{Hyperspectral unmixing} 
Raman spectra can be represented as vectors $\mathbf{x} \in \mathbb{R}_{+}^{b}$, whose components correspond to the intensity of inelastically scattered light binned over $b$ wavelength/wavenumber bands. Any such measurement can be treated as the result of an underlying mixing of $n$ `pure' components, defined by their Raman signatures (endmembers) $ \mathbf{m}_i \in \mathbb{R}_+^{b},  \, i=1,\ldots,n$, and their respective proportions (fractional abundances) $\{ \alpha_i \}_{i=1}^n, \, \alpha_i \in \mathbb{R}_{+}$. Hyperspectral unmixing is the inverse problem of recovering the endmembers and fractional abundances from a measurement $\mathbf{x}$. The unmixing can be performed with respect to a set of known endmembers (non-blind unmixing) or without knowing the endmembers (blind unmixing). Here, we focus on blind unmixing but we also discuss how to adapt the framework to the simpler problem of non-blind unmixing.

A major hurdle for unmixing is the lack of information about the underlying mixing function. The simplest and most common model is the linear mixing model (LMM), where measurements are assumed to be a linear combination of the endmembers:
\begin{equation}
\label{eq:linear_mixing}
    \mathbf{x} = M\bm{\alpha} = \sum_{i=1}^{n}\alpha_i\mathbf{m}_i,
\end{equation}
where $M=\begin{bmatrix}\mathbf{m}_1 & \mathbf{m}_2 & \cdots & \mathbf{m}_n\end{bmatrix}$ is an $b\times n$ non-negative matrix containing the $n$ endmember signatures, and $\bm{\alpha}=(\alpha_{1}, \alpha_{2}, \cdots, \alpha_{n})^\mathsf{T}$ is an $n\times 1$ vector storing the corresponding abundances. A random noise term $\bm{\epsilon} \in \mathbb{R}^{b}$ is also often added to Eq.~\ref{eq:linear_mixing} to model stochastic variations. The abundances $\alpha_i$ are constrained to be non-negative (i.e., the abundance non-negativity constraint (ANC), $\alpha_i \geq 0, \forall i$), and are forced to sum to 1 when corresponding to proportions (i.e., the abundance sum-to-one constraint (ASC), $||\bm{\alpha}||_1 = 1$ ). 

The linear mixing simplification allows the development of tractable approaches for unmixing, such as N-FINDR and VCA for endmember identification, and NNLS and FCLS for abundance estimation. N-FINDR and VCA are geometric methods based on the concept of a simplex in Euclidean space. N-FINDR exploits the fact that, under \eqref{eq:linear_mixing}, endmembers represent vertices of a simplex spanning the data, and operates by iteratively finding a set of points (endmembers) that maximizes the volume of the simplex they form. In contrast, VCA finds endmembers by projecting the data onto directions orthogonal to the subspace spanned by previously found endmembers and identifying new endmembers as the farthest points in these directions, effectively constructing a simplex that encompasses all data points. In both methods, the number of endmembers to extract is specified \textit{a priori} by the user. Once endmember signatures $M$ are derived, optimization-based algorithms such as NNLS and FCLS are employed to estimate the fractional abundances $\bm{\alpha}$ for a given spectrum $\mathbf{x}$ by minimizing the reconstruction error between the observed data and the model $\min_{\bm{\alpha}} \|M\bm{\alpha} - \mathbf{x}\|^2$. NNLS imposes the ANC, whereas FCLS imposes both the ANC and ASC.

The LLM is a good approximation when endmember species are spatially well-separated with respect to the focal volume, and complex light interactions that cause non-linear signal contributions can be neglected~\cite{keshava2002spectral}. However, when non-linear interactions become significant, more intricate models are required~\cite{dobigeon2013nonlinear}. To represent phenomena such as multiple scattering events, topographic variances and shadowing effects, the LMM has been extended in remote sensing to more complex variants~\cite{dobigeon2013nonlinear, heylen2014review}, including intimate mixture models~\cite{hapke1981bidirectional}, bilinear models~\cite{bilinear_models, fan2009comparative, halimi2011unmixing}, multilinear models~\cite{heylen2015multilinear}, or post-nonlinear models~\cite{altmann2012supervised}, among others. 
A popular bilinear mixing model is the Fan model~\cite{fan2009comparative}:
\begin{equation}
\label{eq:bilinear_mixing}
    \mathbf{x} = \sum_{i=1}^{n}\alpha_i\mathbf{m}_i + \sum_{k=1}^{n}\sum_{\substack{l=1, \\ l\neq k}}^{n}\alpha_k\mathbf{m}_k\odot \alpha_l\mathbf{m}_l,
\end{equation}
where $\odot$ is the Hadamard product. However, accounting for non-linear mixing interactions increases the complexity and computational cost of unmixing ~\cite{dobigeon2013nonlinear, heylen2014review}, an issue of especial relevance in Raman spectroscopy where datasets (e.g., imaging/volumetric scans) are typically larger than in remote sensing; hence, despite its limitations, the LMM remains a cornerstone of hyperspectral unmixing in most practical settings~\cite{bioucas2012hyperspectral, li2017spectral}. 

\subsection*{Unmixing autoencoders}
Consider an AE model with an encoder $\mathcal{E}: \mathbb{R}^b \to\mathbb{R}^m$ that transforms spectra $\mathbf{x}$ into a latent space representation $\mathbf{z}=\mathcal{E}(\mathbf{x})$ of some pre-defined dimension $m \ll b$, which is subsequently passed through a decoder $\mathcal{D}: \mathbb{R}^m \to \mathbb{R}^b$ to produce the output reconstruction
\begin{equation}
\label{eq:ae_general}
\widehat{\mathbf{x}}= \mathcal{D}(\mathbf{z}) = \mathcal{D}(\mathcal{E}(\mathbf{x})). 
\end{equation}
Notice that the decoder $\mathcal{D}$ can be understood as playing the role of a mixing function on the representation $\mathbf{z}$. For instance, consider a \emph{linear} decoder $\mathcal{D}_{\text{Lin}}$ consisting of a single linear layer defined by a $b\times m$ weight matrix $W$ then we have
\begin{equation}
\label{eq:ae_linear_mixing}
    \widehat{\mathbf{x}} = \mathcal{D}_{\text{Lin}}(\mathbf{z}) = W\mathbf{z} \, .
\end{equation}
It follows from the formulation of the LMM (Eq.~\ref{eq:linear_mixing}) that the latent representations $\mathbf{z}$ resemble the abundances $\bm{\alpha}$, the weight matrix  $W$ resembles the matrix of endmembers $M$, and the dimensionality $m$ of the latent space defines the number $n$ of endmembers to learn.   

To reinforce the physical interpretation of unmixing into the AE learning, we enforce relevant constraints, such as the non-negativity of $W$ and the non-negativity (ANC) and sum-to-one constraint (ANC) of $\mathbf{z}$, by applying appropriate choices of activation functions and penalties during training.

This framework can accommodate non-linear mixture models through the design of the decoder. For instance, the bilinear Fan model (Eq.~\ref{eq:bilinear_mixing}) can be implemented by extending $\mathcal{D}_{\text{Lin}}$ to account for the additional bilinear terms:
\begin{align}
\label{eq:fan_autoencoder}
    \widehat{\mathbf{x}} = \mathcal{D}_{\text{Bilin}}(\mathbf{z}) = W\mathbf{z} + \sum_{k=1}^{m}\sum_{\substack{l=1, \\ l\neq k}}^{m}z_k \mathbf{w}_k\odot z_l \mathbf{w}_l,
\end{align}
where $z_k, z_l$ are components of $\mathbf{z}$, and $\mathbf{w}_k, \mathbf{w}_l$ are column vectors of $W$. Similarly, one can devise decoders suited for other mixture models~\cite{chen2022integration, shahid2021unsupervised, zhao2021hyperspectral}, or adopt a general decoder that learns the underlying mixing model in a more data-driven manner, at the cost of interpretability of the extracted endmembers and fractional abundances. Finally, note that the AE unmixing framework can be directly adapted for non-blind unmixing by fixing the weight matrix $W$ in the decoder to a given set of predefined endmembers.

\subsection*{Autoencoder architectures}

\subsubsection*{Dense AE} 
This autoencoder employs an encoder comprising $2$ fully connected (or dense) layers.
The first layer projects spectra of dimension $b$ to hidden features of dimension $128$ (Leaky ReLU activation with a slope of $0.02$), which the second layer further projects to latent representations of dimension $n$ ($n$ is the number of endmembers to extract). In the \emph{Deep Dense AE} model used in the analysis of the THP-1 cell, we increase the number of hidden layers to five, comprising $512$, $256$, $128$, $64$ and $32$ neurons, respectively, before the final layer of size $n$.

\subsubsection*{Convolutional AE} 
This model extends the \emph{Dense AE} by adding a convolutional block before the dense layers. The convolutional block consists of two layers of 1D convolutions connected in parallel, each comprising 16 filters of size 3 and 16 filters of size 5 (ReLU activation; input padded with zeroes). The outputs from these two layers are concatenated and merged (channel-wise) via a 2-dimensional dense layer to produce representations of dimension $b$, which are then fed to the \emph{Dense} encoder described above.

\subsubsection*{Transformer AE} 
In this transformer-based encoder, input spectra are first projected to features of size $32$ through a fully connected layer, and then fed to a transformer encoder layer comprising a multi-head attention block with $2$ attention heads of size $32$ \cite{vaswani2017attention}, followed by two fully connected layers expanding the features to size $64$ (ReLU activation) and condensing back to $32$ (no activation). We apply layer normalization \cite{ba2016layer} and dropout ($10\%$)~\cite{srivastava2014dropout} after the multi-head attention block and the fully connected layers. The output of the transformer block is then channeled into the last fully connected layer of size $n$.

\subsubsection*{Convolutional Transformer AE} 
In this model, the \emph{Transformer AE} architecture is extended with the same convolutional block used in the \emph{Convolutional AE}, here added before the transformer-based encoder block.

\subsubsection*{Decoder choice} Our linear unmixing decoder architecture consists of a single fully connected layer using the identity activation function without bias (Eq.~\ref{eq:ae_linear_mixing}). Our bilinear Fan decoder has the same architecture as the linear decoder but also calculates the additional bilinear interaction terms in line with Eq.~\ref{eq:fan_autoencoder}.

\subsubsection*{Physics-inspired constraints}
Fractional abundance constraints are applied through the choice of latent space activation functions. To enforce both  ANC and ASC, we apply a softmax activation function in the final layer in each encoder. When only ANC is used, the activation function is changed to a `softly-rectified' hyperbolic tangent function given by $\frac{1}{\gamma}log(1+e^{\gamma*tanh(x)})$, with $\gamma=10$, which we design to ensure abundances are between $0$ and $1$ but do not necessarily add up to one. To ensure the non-negativity of endmembers, we constrain the weight matrix of the linear layer in the decoder by clipping negative values to zero during training. 

\subsection*{Generating synthetic Raman mixtures}

\subsubsection*{Generating endmembers} For each synthetic dataset, we first generate $n$ endmembers spanning $b$ spectral bands. For the scope of this work, $n=5$ and $b=1000$. Each endmember  $\mathbf{m}_i \in \mathbb{R}_{+}^{b}$ is created by a superposition of a set of $n_{\text{peaks}, i}$ Gaussian peaks of different amplitude, width and location, randomly sampled as follows. The number of peaks is sampled from a discrete uniform distribution $n_{\text{peaks}, i} \sim \mathcal{U} (5,9)$. Each peak $p$ is described by $p = h_p \sigma_p \sqrt{2\pi} \, \mathcal{N}(b_p, \sigma_p)$, where $\mathcal{N}(\cdot)$ represents a Gaussian distribution. The height of the peak is defined as $h_p=h_1 \cdot h_2$, where $h_1=1+5 \, h_\beta$ with $h_\beta\sim\text{Beta}(1, 3)$ and $h_2\sim \mathcal{U}(0.1, 1)$. The center of the peak is sampled from $b_p \sim \mathcal{U} (10, b-10)$, and the width of the peak is defined as $\sigma_p=w_p \sigma$, with $\sigma\sim \mathcal{U}(0.1, 1)$. 

We create two types of endmembers: \emph{clean} and \emph{noisy}. For the former, we produce peaks with  $w_p=1$. For the latter, we augment \emph{clean} endmembers by adding $n_{\text{peaks}, i}^{\text{small}} \sim \mathcal{U} (50,99)$ smaller peaks sampled with $h_1=1/3$ and $w_p=2$, thus making \emph{noisy} endmembers better resemble experimental Raman signatures.
 
\subsubsection*{Generating fractional abundances} 

For visualization purposes, we present the fractional abundance profiles in the form of two-dimensional scenes comprising $H\times W$ pixels, where each pixel represents a fractional abundance vector $\bm{\alpha}\in \mathbb{R}_{+}^{n}$. Here, we set $H=W=100$, resulting in $10000$ spectra per scene/dataset. In the simplest scene (\emph{Chessboard}), we split the scene into $20 \times 20$ square patches, each containing a single randomly assigned endmember (i.e., all $400$ pixels in each patch are the same one-hot vector). Our second scene (\emph{Gaussian}) consists of $n$ Gaussian functions equally spaced along the diagonal of the scene. After each pixel is normalized to comply with the ASC, we obtain abundance profiles representing different levels of overlap of components. Our last fractional abundance scene (\emph{Dirichlet}) corresponds to a highly mixed scene, where each pixel is individually sampled from a $n$-dimensional Dirichlet distribution, producing a random mixture of all endmembers. 
Note that the fractional abundance profile of each pixel in all three scenes complies with both ANC and ASC.  

\subsubsection*{Mixing model}  

Having generated a set of endmembers and an underlying fractional abundance scene, mixed data measurements $\mathbf{x}\in \mathbb{R}^{b}$ are created based on a mixing model chosen by the user. In this study, we consider linear mixtures (Eq.~\ref{eq:linear_mixing}) and bilinear mixtures based on the Fan model (Eq.~\ref{eq:bilinear_mixing}).

\subsubsection*{Adding data artifacts} Finally, data artifacts (noise, baseline, cosmic spikes) can be optionally added to create more realistic synthetic Raman spectra. Here, we add Gaussian noise $\bm{\epsilon} \in \mathbb{R}^{b}$ to each spectrum, with independent and identically distributed components $\mathbf{\epsilon}_i\sim\mathcal{N}(0, \sigma_N)$. Further, we add a baseline signal $\mathbf{B}= h_B \arctan(\pi [1:b]/b) \in \mathbb{R}^{b} $ to each spectrum with probability $p_B$. Finally, with probability $p_S$, a cosmic spike of intensity $S \sim h_S U(0.75, 1.25)$ is added to each spectrum at a band $b_S \sim\mathcal{U}\{2, b-2\}$. In our experiments: $\sigma_N=0.1$, $p_B=0.25$ $h_B=2$, , $p_S=0.1$, $h_S=5$.

\subsection*{Model training and evaluation on synthetic Raman mixtures}

Autoencoders were trained on synthetic data using the Adam optimizer (learning rate $0.001$) over $10$ epochs, with spectral angle distance (SAD)~\cite{sad}, see Eq.~\ref{eq:SAD_def},  as a loss function between input and reconstructed spectra. The latent dimensionality $m$ of each AE model is set to $5$ for the \emph{ideal} mixture scenario, and $6$ for the other mixture scenarios with data artifacts. Both ANC and ASC are enforced for all experiments on synthetic data.

To quantify the accuracy, we utilize two measures. The MSE measures the distance between ground truth  and predicted fractional abundances $\bm{\alpha}$ and $\widehat{\bm{\alpha}}$:
\begin{equation}
\label{eq:MSE_def}
    \text{MSE}(\bm{\alpha}, \widehat{\bm{\alpha}}) = 
     \frac{1}{n} ||\bm{\alpha}-\widehat{\bm{\alpha}}||_2.
\end{equation} 
The scale-invariant SAD measures the discrepancy between ground truth and predicted endmembers $\mathbf{m}_i$ and $\widehat{\mathbf{m}}_i$:
\begin{equation}
\label{eq:SAD_def}
    0 \leq \text{SAD}(\mathbf{m}_i, \widehat{\mathbf{m}}_i) = \arccos{\left(\frac{\mathbf{m}_i \cdot \widehat{\mathbf{m}}_i}
    {\Vert \mathbf{m}_i\Vert_2 \, \Vert \widehat{\mathbf{m}}_i\Vert_2}\right)} \leq 1.
\end{equation}

For each evaluation, we first assign the derived and ground truth endmembers (and corresponding fractional abundances) via the Hungarian algorithm with SAD as the objective to minimize. When the number of extracted endmembers $n$ is higher than the number of ground truth endmembers $n_{\text{true}}$, we only use $n_{\text{true}}$ endmember and corresponding fractional abundance estimates to compute the performance metrics.

Each experiment on the synthetic data was performed on $5$ datasets and $5$ model initializations using different random seeds, resulting in $5\times 5=25$ replicates per evaluation, or $1650$ experiments in total: $6$ models (2 conventional, 4 AEs) $\times$ $11$ dataset variants $\times$ $25$ replicates. Random seeds were kept the same across mixture scenarios to allow direct comparison. 

\subsection*{Measuring computational cost}
We profile the computational cost of unmixing methods on synthetic datasets (\emph{ideal} scenario, \emph{Chessboard} scene) of increasing sizes, from $2500$ to $250000$ spectra. The number of endmembers to extract was set to $n=5$ for all methods. For each experiment, we performed $3$ separate evaluations, measuring the wall time of each method (including the training time for autoencoders). All experiments were conducted on a MacBook Air laptop with an Apple M2 chip (8-core CPU, 10-core GPU, and 16-core Neural Engine). We only employed CPU computations to ensure a fair comparison with traditional methods which, by design,  do not utilize GPU acceleration. 

\subsection*{Analysis of experimental RS data from sugar mixtures}
\subsubsection*{Preparation of sugar solutions}
We prepared \qty[per-mode = symbol]{1}{\mol\per\litre} solutions of each type of sugar (sucrose, fructose, maltose, and glucose) by dissolving the appropriate weight of sugar into \qty{40}{\ml} of ultrapure distilled water (Invitrogen™ -- UltraPure™ DNase/RNase-Free Distilled Water). The weights of sugars dissolved were \qty{13.83}{\gram} for sucrose (Thermo Scientific Chemicals -- Sucrose, $99$\%), \qty{7.279}{\gram} for fructose (Thermo Scientific Chemicals -- D-Fructose, $99$\%), \qty{15.171}{\gram} for maltose (Thermo Scientific Chemicals -- D-(+)-Maltose monohydrate, $95$\%) and \qty{7.279}{\gram} for glucose (D-(+)-Glucose, AnalaR NORMAPUR$^{\text{\textregistered}}$ analytical reagent). All solutions were mixed and vortexed in standard \qty{50}{\ml} centrifuge tubes until no solute was visible.

Sugar mixtures were prepared in standard 96-well plates, with a volume of \qty{375}{\ul} per well. A full factorial experiment was performed comprising $4$ volume levels for each sugar (\qtylist{0;30;75;120}{\ul}), filled with distilled water where necessary. 
Discarding the mixtures exceeding the volume of the well and the one that contains no sugar, $240$ distinct mixtures were prepared. 
In addition, $5$ extra `pure' solutions (i.e., \qty{375}{\ul} of water, sucrose, fructose, maltose, or glucose) were prepared, which we used to extract reference spectra for each chemical species. This resulted in a total of $245$ wells distributed in three standard $96$-well plates. Mixtures were stirred using standard \qty{200}{\ul} pipettes before spectral acquisition to ensure good mixing.

\subsubsection*{Raman measurements from sugar solutions}
All spectra were acquired using a custom Raman microspectroscopy platform designed for high-throughput analysis known as B-Raman. This platform is based on the Thorlabs Cerna$^{\text{\textregistered}}$ and features the BWTek BRM-785-0.55-100-0.22-SMA laser excitation source and the Ibsen EAGLE Raman-S spectrometer. The instrument was calibrated using an Argon wavelength calibration source (AR-2 -- Ocean Insight) reference lamp before data collection. The excitation wavelength was \qty{785}{\nm} and the power incident to the samples was \qty{36.3}{\mW}. The Raman scattering was collected in reflection via a Leica N PLAN 10x/0.25 objective with $0.25$ numerical aperture. The raw spectra were acquired over the spectral wavenumber range of \numrange[range-phrase=--]{142}{3684.8}\,\unit{\per\centi\metre}.

Spectra were measured from the center (horizontal) of each well at a fixed depth that was established to provide the highest signal. Two sets of data were collected from each well, at \qty{5}{\s} and \qty{0.5}{\s} integration times, to compare unmixing performance on low and high signal-to-noise ratio (SNR) data. Several measurements were collected from each well, resulting in a total of $240\text{ solutions}\times2\text{ measurements}\times4\text{ repetitions}=1920$ high-SNR measurements ($1960$ with reference spectra); and $240\text{ solutions}\times8\text{ measurements}\times4\text{ repetitions}=7680$ low-SNR measurements ($7840$ with reference spectra). Ground-truth endmembers signatures were obtained by taking the median (band-wise) of the reference spectra ($40$ in high SNR setup, and $160$ in low SNR setup) collected from the $5$ additional wells containing pure solutions. Ground truth fractional abundances were determined by calculating the ratio of the components present in each mixture. 

\subsubsection*{Preprocessing and analysis of sugar data}
First, we preprocess each sugar dataset: 1) cropping to the region \numrange[range-phrase=--]{400}{1800}\,\unit{\per\centi\metre}; 2) baseline correction with Adaptive Smoothness Parameter Penalized Least Squares (ASPLS)~\cite{zhang2020baseline}---smoothing parameter $\lambda=10^5$, differential matrix of order $2$, maximum iterations set to $100$, exit criteria with tolerance $t=0.001$; 3) global vector normalization, where each observation is divided by the highest magnitude observed in the data. Baseline removal is important to ensure models extract relevant features (i.e., characteristic peaks) as opposed to merely capturing the trend. 

To perform hyperspectral unmixing, we set the number of endmembers to extract to $n=5$, and we follow similar training and evaluation protocols to those employed for the synthetic data, but we increase the number of epochs to $15$ for low SNR data and $50$ for high SNR data given the more limited number of spectra collected. We also incorporate an additional MSE term in the training loss $\mathcal{L}$ of autoencoders on high SNR data:  
\begin{equation}
    \mathcal{L}(\mathbf{x}, \widehat{\mathbf{x}})=\text{SAD}(\mathbf{x}, \widehat{\mathbf{x}})+\lambda \, \text{MSE}(\mathbf{x}, \widehat{\mathbf{x}}),
\end{equation}
with $\lambda=1000$. This term breaks the invariance to scale and leads to better abundance estimation given the weak water endmember (see Table~\ref{tab:sugar_metrics_full}). The standard SAD loss was used for experiments on low SNR. Each experiment is repeated for $5$ model initializations.

Table~\ref{tab:sugar_metrics_ncc} presents performance evaluation using an alternative endmember distance based on Pearson’s correlation coefficient (PCC), showing an even more pronounced improvement in endmember estimation accuracy.

\subsection*{Analysis of volumetric RS data from THP-1 cell}
The volumetric Raman scan of the THP-1 cell~\cite{kallepitis2017quantitative} was collected using \qty{0.3}{\s} integration time and comprises a $z=1,\ldots, 10$ stack of ten $40\times 40$ raster scans, organized into a single volumetric hypercube for analysis. We preprocess the data before unmixing using the following protocol: 1) spectral cropping to the fingerprint region \numrange[range-phrase=--]{700}{1800}\,\unit{\per\centi\metre}; 2) cosmic spike removal using the algorithm in~\cite{whitaker2018simple} with kernel of size $3$ and z-value threshold of $8$; 3) denoising with Savitzky-Golay filter using a cubic polynomial kernel of size  $7$~\cite{savgol}; 4) baseline correction using Asymmetric Least Squares (AsLS) with smoothing parameter $\lambda=10^6$, penalizing weighting factor $p=0.01$, differential matrix of order $2$, maximum iterations set to $50$, exit criteria with tolerance threshold of $t=0.001$~\cite{eilers2005baseline}; 5) global MinMax normalization to the interval $[0, 1]$.

Unmixing is performed following the same AE training protocol as in other analyses, with the number of training epochs set to $20$, and the number of endmembers to extract to $n=20$. Here, we also discard the constraint that fractional abundances must sum to one. Out of the $20$ endmembers we obtain, we display the $5$ deemed most biologically relevant following peak assignment as per the original paper~\cite{kallepitis2017quantitative}. For VCA+NNLS, two of those five endmembers corresponded to the same cell organelle, namely cytoplasm, and were visualized using the same color in the merged reconstruction displayed in Fig.~\ref{fig:real_application}b.

Cell organelles were determined based on the following peaks: PBS buffer - \qty{1637}{\per\centi\metre} (water peak); cytoplasm - \qty{1005}{\per\centi\metre} (phenylalanine), \qty{1250}{\per\centi\metre} (Amide III), \qty{1659}{\per\centi\metre} (Amide I) and \qty{1445}{\per\centi\metre} (CH deformations of proteins and lipids); TAGs/PLPs - \qty{1092}{\per\centi\metre} (C–C stretching ), \qty{1308}{\per\centi\metre} ($\text{CH}_2$ twists), \qty{1445}{\per\centi\metre} (CH deformation), and \qty{1661}{\per\centi\metre} (C=C stretching); nucleus/DNA - \qty{790}{\per\centi\metre} (symmetric phosphodiester stretch and ring breathing modes of pyrimidine bases) and \qty{1103}{\per\centi\metre} (symmetric dioxy-stretch of the phosphate backbone); cholesterol - \qty{1069}{\per\centi\metre} and \qty{1134}{\per\centi\metre} (cholesteryl stearate), \qty{1300}{\per\centi\metre} ($\text{CH}_2$ twists), and \qty{1443}{\per\centi\metre} (CH deformation)~\cite{kallepitis2017quantitative, zhang2012label, movasaghi2007raman}. 

\end{doublespacing}

\section{Implementation}
We conduct our analyses in Python, using \emph{tensorflow}~\cite{tensorflow} for autoencoder model development and training, and the \emph{RamanSPy} package~\cite{georgiev2023ramanspy} for unmixing with conventional methods, data loading and management, preprocessing, and plotting.

\section{Data availability}
Our synthetic data generator has been integrated into the \emph{RamanSPy} package (\url{https://github.com/barahona-research-group/RamanSPy})~\cite{georgiev2023ramanspy}. THP-1 cell data has been previously deposited by the authors of the original paper at \url{https://zenodo.org/record/256329} (scan \textit{'001'}). The rest of our experimental data will be made available upon publication.

\section{Acknowledgments}
D.G. and G.P. are supported by UK Research and Innovation [UKRI Centre for Doctoral Training in AI for Healthcare grant number EP/S023283/1].  
A.F.G. acknowledges support from the Schmidt Science Fellows, in partnership with the Rhodes Trust.
S.V.P. acknowledges support from the Independent Research Fund Denmark (0170-00011B).
R.X. and M.M.S. acknowledge support from the Engineering and Physical Sciences Research Council (EP/P00114/1 and EP/T020792/1).
M.M.S. acknowledges support from the Royal Academy of Engineering Chair in Emerging Technologies award (CiET2021\textbackslash\textbackslash94) and the Bio Innovation Institute.
M.B. acknowledges support from the Engineering and Physical Sciences Research Council (EP/N014529/1, funding the EPSRC Centre for Mathematics of Precision Healthcare at Imperial, and EP/T027258/1).
The authors thank Dr Akemi Nogiwa Valdez for proofreading and data management support. 
Figures were assembled in BioRender.

\section{Author Contributions}
D.G. conceived the project, implemented software, and conducted experiments and analyses. M.B., M.M.S., A.F.G. and S.V.P. contributed to conceptualization and supervised the study. S.V.P., A.F.G. and M.B. contributed to methods and data review, and interpretation of results. A.F.G. and G.P. prepared sugar solutions, with A.F.G. performing consecutive data acquisition. R.X. contributed to interpretation of cell analysis. D.G. wrote the original draft, and revised it together with M.B., with input from A.F.G., S.V.P., R.X. and M.M.S. All authors have approved the final manuscript.

\section{Author Competing Interests}
M.M.S. invested in, consults for (or was on scientific advisory boards or boards of directors) and conducts sponsored research funded by companies related to the biomaterials field. M.M.S. holds a grant from the BII program in Copenhagen which supports project ASAI around potential future commercialisation of SERS-based biosensing. The rest of the authors declare no competing interests.

\renewcommand\refname{References}
\begin{footnotesize}
\bibliographystyle{unsrt.bst} 
\textnormal{\bibliography{main.bbl}}
\end{footnotesize}

\newpage

\bgroup\setlength{\parindent}{0pt}
\begin{flushleft}
    \begin{adjustwidth}{0in}{.5in} 
        \begin{Huge} 
            \centering
            \begin{spacing}{.9} 
            \textbf{Supplementary Information}
            \end{spacing}
        \end{Huge}

        \vspace{1.5cm}

        \begin{huge} 
            \begin{spacing}{.9} 
            \textbf{Hyperspectral unmixing for Raman spectroscopy via physics-constrained autoencoders}
            \end{spacing}
        \end{huge}

        \vspace{0.5cm}

        \begin{large} 
            \begin{spacing}{.9} 
            \textbf{Dimitar Georgiev, Álvaro Fernández-Galiana, Simon Vilms Pedersen, Georgios Papadopoulos, Ruoxiao Xie, Molly M. Stevens\textsuperscript{\#}, Mauricio Barahona\textsuperscript{\#}}\\[0.25cm]
            \textsuperscript{\#}E-mail: molly.stevens@dpag.ox.ac.uk \& m.barahona@imperial.ac.uk
            \end{spacing}
        \end{large}

    \end{adjustwidth}
\end{flushleft}
\egroup

\vspace{5cm}

\beginsupplement 

\begin{table*}[!h]
    \footnotesize
    \centering
    \begin{tabular}{p{4cm}||p{5cm}|p{6cm}}
    \textbf{} & \textbf{Commonly used methods} & \textbf{Autoencoders} \\
    \hline
    \hline
    Assumptions & Strict - e.g. endmembers present in the data; fixed mixture model; number of endmembers is known. &  Autoencoders can facilitate data-driven unmixing without many strong assumptions about the data and task.\\\hline
    Mixture model & A specific model is assumed \textit{a priori}. Usually, a linear mixing only, as alternatives can become too complex and computationally expensive. & Autoencoders can be designed to capture various mixture models, including non-linear mixtures. This can be enforced by appropriate architectural constraints or learned in a data-driven manner. 
    \\\hline
    Input modality \& capturing spatial information & Spectra are processed individually. If imaging or volumetric data are provided, these are typically unfolded and used pixel-wise, thereby discarding any spatial information. & Autoencoders can be extended to utilize the spatial information available in imaging or volumetric data by incorporating (2D, 3D, and/or 4D) convolutional layers. \\\hline
    Simultaneously deriving endmembers and fractional abundances? & Typically, two separate algorithms are applied. & Autoencoders simultaneously extract both endmembers and fractional abundances by default. They can also be adjusted for non-blind unmixing by fixing a pre-defined endmember matrix in the decoder. \\\hline
    Robustness to non-specific signals and preprocessing – e.g. noise, baseline & Performance is usually extremely dependent on the quality of the data, as well as on preprocessing, which can vary greatly from application to application. & Autoencoders can incorporate feature selection blocks and training loss to promote invariance to scaling, baselines, and noise, making them more robust to artifacts and outliers. \\\hline
    Variable spectral axis & Observations are assumed to share a common spectral axis, and the spectral axis is discarded. This impedes the integration of data across experimental setups unless they share the same spectral axis. & The axis can be integrated as an input to the model (e.g. as a positional encoding), allowing the direct integration of diverse data sources. \\\hline
    Scalability & Even the simplest models can become computationally prohibitive for larger datasets. More advanced options are intractable in real-world applications involving imaging or volumetric Raman scans. & Deep neural networks are designed to be scalable and parallelizable out-of-the-box. \\\hline
    Number of endmembers & Typically, fixed \textit{a priori}. & Normally fixed \textit{a priori}, too, but can potentially be learned by introducing sparsity and information-theoretic criteria in the latent space, for instance. \\\hline
    Availability & Most algorithms are not widely available in common open-source programming environments, such as Python. & With the availability of modern deep learning libraries and frameworks, implementing and integrating autoencoders into existing workflows and pipelines has become widely available. \\\hline
    Extensibility & Most conventional techniques are based on specialized optimization-based algorithms tailored for the particular unmixing task (e.g. mixture model, scene type, number of endmembers), which makes their extension challenging. & Easy extension and adaptation. One can readily expand unmixing autoencoders into more advanced variants, as well as integrate models into established AI and ML pipelines.  \\\hline
    Downstream applications & Conventional techniques are generally constrained within hyperspectral unmixing, with algorithms only outputting the derived unmixing results. & Autoencoders can be used as feature extractors in a variety of modeling and predictive downstream tasks, such as classification, clustering and anomaly detection. They can also leverage transfer learning techniques to improve performance in situations with limited labeled data by utilizing unmixing autoencoders for self-supervised pre-training. \\\hline
    \end{tabular}
    \caption{
    \textbf{Advantages of autoencoders over standard unmixing techniques.} 
    {\normalfont Autoencoders provide a more flexible framework for hyperspectral unmixing, offering a multitude of advantages over traditional methods for unmixing.} }   
    \label{tab:advantages}
\end{table*}

\begin{table*}[b]
    \centering
    \tiny
    \begin{tabular}{llcccccccc}
    \toprule
    &  & \multicolumn{2}{c}{\emph{Chessboard}} & & \multicolumn{2}{c}{\emph{Gaussian}} & & \multicolumn{2}{c}{\emph{Dirichlet}} 
    \\\cmidrule(r){3-4}\cmidrule(r){5-7}\cmidrule(r){8-10}    
    &  & Endmembers & Abundances & &  Endmembers & Abundances &  & Endmembers & Abundances\\
    & Method  & (SAD) $\downarrow$ & (MSE) $\downarrow$ &  & (SAD) $\downarrow$ & (MSE) $\downarrow$ &  & (SAD) $\downarrow$ & (MSE) $\downarrow$\\
    \toprule
    \parbox[t]{1mm}{\multirow{7}{*}{\rotatebox[origin=c]{90}{\emph{ideal}}}} 
    & PCA                                  & 0.874 $\pm$ 0.139                & 22.592 $\pm$ 8.969               &  & 1.051 $\pm$ 0.081                        & 5.387 $\pm$ 1.876                &  & 0.847 $\pm$ 0.092                        & 3.811 $\pm$ 1.291                \\
    & N-FINDR + FCLS                      & \underline{\textbf{0.000}} $\pm$ \underline{\textbf{0.000}}     & \underline{\textbf{0.000}} $\pm$ \underline{\textbf{0.000}}     &  & 0.473 $\pm$ 0.042                        & 0.033 $\pm$ 0.001                &  & 0.101 $\pm$ 0.040                         & \underline{\textbf{0.001}} $\pm$ \underline{\textbf{0.000}}   \\
    & VCA + FCLS                             & \underline{\textbf{0.000}} $\pm$ \underline{\textbf{0.000}}     & \underline{\textbf{0.000}} $\pm$ \underline{\textbf{0.000}}     &  & 0.482 $\pm$ 0.047                        & 0.038 $\pm$ 0.009                &  & 0.105 $\pm$ 0.031                        & \textbf{0.003} $\pm$ \textbf{0.003}       \\
    \cmidrule(r){2-10}
    & Dense AE                              & 0.015 $\pm$ 0.065                & 0.002 $\pm$ 0.008                &  & 0.255 $\pm$ 0.037                        & \textbf{0.010} $\pm$ \textbf{0.005}        &  & \underline{\textbf{0.003}} $\pm$ \underline{\textbf{0.001}}         & 0.012 $\pm$ 0.007                \\
    & Convolutional AE                       & 0.042 $\pm$ 0.111                & 0.002 $\pm$ 0.007                &  & \underline{\textbf{0.168}} $\pm$ \underline{\textbf{0.033}}         & \underline{\textbf{0.009}} $\pm$ \underline{\textbf{0.006}} &  & \underline{\textbf{0.003}} $\pm$ \underline{\textbf{0.001}}         & 0.012 $\pm$ 0.006                \\
    & Transformer AE                         & \textbf{0.002} $\pm$ \textbf{0.000}         & \underline{\textbf{0.000}} $\pm$ \underline{\textbf{0.000}}     &  & \textbf{0.212} $\pm$ \textbf{0.043}               & \underline{\textbf{0.009}} $\pm$ \underline{\textbf{0.005}} &  & \textbf{0.007} $\pm$ \textbf{0.002}               & 0.012 $\pm$ 0.006                \\
    & Convolutional Transformer AE            & \textbf{0.002} $\pm$ \textbf{0.000}         & \underline{\textbf{0.000}} $\pm$ \underline{\textbf{0.000}}     &  & 0.215 $\pm$ 0.042                        & \underline{\textbf{0.009}} $\pm$ \underline{\textbf{0.005}} &  & 0.017 $\pm$ 0.062                        & 0.013 $\pm$ 0.007                \\
    \cmidrule(r){1-10}\morecmidrules\cmidrule(r){1-10}
    \parbox[t]{1mm}{\multirow{7}{*}{\rotatebox[origin=c]{90}{\emph{+ artifacts}}}}
    & PCA                                     & 0.819 $\pm$ 0.072                & 79.38 $\pm$ 49.727               &  & 1.009 $\pm$ 0.082                        & 61.488 $\pm$ 48.586              &  & 0.869 $\pm$ 0.069                        & 60.154 $\pm$ 48.330               \\
    & N-FINDR + FCLS                          & 0.629 $\pm$ 0.111                & 0.073 $\pm$ 0.036                &  & 0.828 $\pm$ 0.075                        & 0.041 $\pm$ 0.015                &  & 0.599 $\pm$ 0.116                        & 0.033 $\pm$ 0.015                \\
    & VCA + FCLS                               & 0.353 $\pm$ 0.159                & 0.071 $\pm$ 0.042                &  & 0.609 $\pm$ 0.093                        & 0.059 $\pm$ 0.015                &  & 0.392 $\pm$ 0.175                        & 0.036 $\pm$ 0.019                \\
    \cmidrule(r){2-10}
    & Dense AE                               & 0.072 $\pm$ 0.081                & 0.042 $\pm$ 0.032                &  & \underline{\textbf{0.347}} $\pm$ \underline{\textbf{0.137}}         & 0.029 $\pm$ 0.021                &  & \underline{\textbf{0.068}} $\pm$ \underline{\textbf{0.070}}          & 0.021 $\pm$ 0.011                \\
    & Convolutional AE                         & \underline{\textbf{0.033}} $\pm$ \underline{\textbf{0.003}} & 0.023 $\pm$ 0.013                &  & \textbf{0.385} $\pm$ \textbf{0.166}               & \textbf{0.027} $\pm$ \textbf{0.014}       &  & \underline{\textbf{0.068}} $\pm$ \underline{\textbf{0.072}}         & \underline{\textbf{0.017}} $\pm$ \underline{\textbf{0.008}} \\
    & Transformer AE                        & \textbf{0.039} $\pm$ \textbf{0.031}       & \textbf{0.022} $\pm$ \textbf{0.011}       &  & 0.393 $\pm$ 0.129                        & \underline{\textbf{0.026}} $\pm$ \underline{\textbf{0.014}} &  & \textbf{0.073} $\pm$ \textbf{0.105}               & \underline{\textbf{0.017}} $\pm$ \underline{\textbf{0.009}} \\
    & Convolutional Transformer AE           & \underline{\textbf{0.033}} $\pm$ \underline{\textbf{0.004}} & \underline{\textbf{0.021}} $\pm$ \underline{\textbf{0.010}}  &  & 0.399 $\pm$ 0.164                        & \textbf{0.027} $\pm$ \textbf{0.015}       &  & 0.112 $\pm$ 0.140                         & \textbf{0.019} $\pm$ \textbf{0.010}        \\
    \cmidrule(r){1-10}\morecmidrules\cmidrule(r){1-10}
    \parbox[t]{1mm}{\multirow{7}{*}{\rotatebox[origin=c]{90}{\emph{+ realistic}}}}
    & PCA                                  & 0.967 $\pm$ 0.074                & 88.132 $\pm$ 50.053              &  & 1.079 $\pm$ 0.045                        & 63.735 $\pm$ 48.513              &  & 0.993 $\pm$ 0.089                        & 61.244 $\pm$ 48.733              \\
    & N-FINDR + FCLS                         & 0.361 $\pm$ 0.062                & 0.072 $\pm$ 0.040                 &  & 0.478 $\pm$ 0.047                        & 0.043 $\pm$ 0.013                &  & 0.299 $\pm$ 0.087                        & 0.030 $\pm$ 0.019                 \\
    & VCA + FCLS                              & 0.173 $\pm$ 0.079                & 0.061 $\pm$ 0.038                &  & 0.400 $\pm$ 0.075                          & 0.063 $\pm$ 0.016                &  & 0.229 $\pm$ 0.088                        & 0.030 $\pm$ 0.017                 \\
    \cmidrule(r){2-10}
    & Dense AE                               & 0.045 $\pm$ 0.024                & 0.030 $\pm$ 0.023                 &  & 0.177 $\pm$ 0.050                         & 0.015 $\pm$ 0.011                &  & \textbf{0.073} $\pm$ \textbf{0.012}               & \textbf{0.011} $\pm$ \textbf{0.006}       \\
    & Convolutional AE                       & 0.054 $\pm$ 0.013                & \textbf{0.019} $\pm$ \textbf{0.008}       &  & 0.168 $\pm$ 0.030                         & \textbf{0.011} $\pm$ \textbf{0.004}       &  & \underline{\textbf{0.071}} $\pm$ \underline{\textbf{0.022}}         & \underline{\textbf{0.008}} $\pm$ \underline{\textbf{0.003}} \\
    & Transformer AE                      & \underline{\textbf{0.039}} $\pm$ \underline{\textbf{0.005}} & \underline{\textbf{0.018}} $\pm$ \underline{\textbf{0.009}} &  & \textbf{0.156} $\pm$ \textbf{0.044}               & \textbf{0.011} $\pm$ \textbf{0.005}       &  & 0.085 $\pm$ 0.014                        & \underline{\textbf{0.008}} $\pm$ \underline{\textbf{0.003}} \\
    & Convolutional Transformer AE           & \textbf{0.040} $\pm$ \textbf{0.005}        & \underline{\textbf{0.018}} $\pm$ \underline{\textbf{0.008}} &  & \underline{\textbf{0.151}} $\pm$ \underline{\textbf{0.033}}         & \underline{\textbf{0.010}} $\pm$ \underline{\textbf{0.004}}  &  & 0.081 $\pm$ 0.012                        & \underline{\textbf{0.008}} $\pm$ \underline{\textbf{0.003}} \\
    \cmidrule(r){1-10}\morecmidrules\cmidrule(r){1-10}
    \parbox[t]{1mm}{\multirow{7}{*}{\rotatebox[origin=c]{90}{\emph{+ bilinear}}}}
    & PCA                                &\dittotikz                    &\dittotikz                    &  & 1.117 $\pm$ 0.071                        & 67.189 $\pm$ 48.404              &  & 1.038 $\pm$ 0.112                        & 63.915 $\pm$ 48.423              \\
    & N-FINDR + FCLS                     &\dittotikz                    &\dittotikz                    &  & 0.456 $\pm$ 0.049                        & 0.039 $\pm$ 0.016                &  & 0.287 $\pm$ 0.074                        & 0.025 $\pm$ 0.015                \\
    & VCA + FCLS                           &\dittotikz                    &\dittotikz                    &  & 0.391 $\pm$ 0.058                        & 0.057 $\pm$ 0.018                &  & 0.277 $\pm$ 0.082                        & 0.030 $\pm$ 0.015                 \\
    \cmidrule(r){2-10}
    & Dense AE (bilinear)             &\dittotikz                    &\dittotikz                    &  & 0.247 $\pm$ 0.070                         & 0.017 $\pm$ 0.011                &  & \textbf{0.094} $\pm$ \textbf{0.011}               & \textbf{0.010} $\pm$ \textbf{0.005}                 \\
    & Convolutional AE (bilinear)*         &\dittotikz                    &\dittotikz                    &  & \underline{\textit{\textbf{0.194}}} $\pm$ \underline{\textbf{0.030}} & \underline{\textbf{0.010}} $\pm$ \underline{\textbf{0.004}}        &  & \underline{\textbf{0.087}} $\pm$ \underline{\textbf{0.010}} & \underline{\textbf{0.008}} $\pm$ \underline{\textbf{0.003}} \\
    & Transformer AE (bilinear)       &\dittotikz                    &\dittotikz                    &  & 0.222 $\pm$ 0.084                        & 0.012 $\pm$ 0.008                &  & 0.105 $\pm$ 0.012                        & \underline{\textbf{0.008 $\pm$ 0.003}} \\
    & Convolutional Transformer AE (bilinear) &\dittotikz                    &\dittotikz                    &  & \textbf{0.208} $\pm$ \textbf{0.051}                        & \textbf{0.011} $\pm$ \textbf{0.006}                &  & 0.100 $\pm$ 0.010                           & \underline{\textbf{0.008 $\pm$ 0.003}} \\
    \bottomrule
    \end{tabular}
    \caption{
    \textbf{Benchmark results on diverse synthetic Raman mixture datasets.} 
    {\normalfont 
    Each value represents the average result of $25$ replicates, including $5$ dataset and $5$ model initialisations. Confidence intervals are calculated as one standard deviation around the sample mean. Values rounded to $3$ decimal places. Best and second best results (mean value) are given in \underline{\textbf{underlined bold}} and \textbf{bold}, respectively.
    }
    }
    \label{tab:synthetic_metrics}
\end{table*}

\begin{table*}[!h]
    \centering
    \tiny
    \begin{tabular}{llcccccccc}
    \toprule
    & & \multicolumn{4}{c}{\emph{High SNR}}& \multicolumn{4}{c}{\emph{Low SNR}}
    \\\cmidrule(r){3-6}\cmidrule(r){7-10}   
    & & \multicolumn{2}{c}{\emph{with endmembers}} & \multicolumn{2}{c}{\emph{without endmembers}} & \multicolumn{2}{c}{\emph{with endmembers}} & \multicolumn{2}{c}{\emph{without endmembers}} 
    \\\cmidrule(r){3-4}\cmidrule(r){5-6}\cmidrule(r){7-8}\cmidrule(r){9-10}  
    & & Endmembers & Abundances & Endmembers & Abundances & Endmembers & Abundances & Endmembers & Abundances\\
    & Method & (SAD) $\downarrow$ & (MSE) $\downarrow$ & (SAD) $\downarrow$ & (MSE) $\downarrow$ & (SAD) $\downarrow$ & (MSE) $\downarrow$ & (SAD) $\downarrow$ & (MSE) $\downarrow$\\
    \toprule
    & PCA                          & 1.018 $\pm$ 0.000                 & 0.073 $\pm$ 0.000                 & 1.026 $\pm$ 0.000                 & 0.072 $\pm$ 0.000                 & 1.173 $\pm$ 0.000                 & 0.075 $\pm$ 0.000               & 1.210 $\pm$ 0.000                  & 0.073 $\pm$ 0.000                 \\
    & N-FINDR+FCLS               & \underline{\textbf{0.202}} $\pm$ \underline{\textbf{0.000}}   & 0.005 $\pm$ 0.000         & 0.515 $\pm$ 0.035                & 0.037 $\pm$ 0.029 & 0.900 $\pm$ 0.000                   & \underline{\textbf{0.026}} $\pm$ \underline{\textbf{0.000}} & 1.077 $\pm$ 0.000                 & 0.154 $\pm$ 0.000                 \\
    & VCA+FCLS                   & 0.322 $\pm$ 0.044                & \textbf{0.004} $\pm$ \textbf{0.000}   & 0.525 $\pm$ 0.026                & 0.042 $\pm$ 0.029       & 0.677 $\pm$ 0.022                & 0.077 $\pm$ 0.035              & 0.901 $\pm$ 0.017                & 0.108 $\pm$ 0.006                \\
    \cmidrule(r){1-10}
    \parbox[t]{1mm}{\multirow{4}{*}{\rotatebox[origin=c]{90}{\emph{SAD}}}} & Dense AE                     & 0.212 $\pm$ 0.004                & 0.057 $\pm$ 0.001                & 0.219 $\pm$ 0.001                 & 0.058 $\pm$ 0.001                & \underline{\textbf{0.462}} $\pm$ \underline{\textbf{0.038}} & \textbf{0.058} $\pm$ \textbf{0.005}     & \underline{\textbf{0.503}} $\pm$ \underline{\textbf{0.046}}       & \underline{\textbf{0.054}} $\pm$ \underline{\textbf{0.007}} \\
    & Convolutional AE             & \textbf{0.203} $\pm$ \textbf{0.001}       & 0.058 $\pm$ 0.001                & \underline{\textbf{0.215}} $\pm$ \underline{\textbf{0.002}} & 0.058 $\pm$ 0.001                & 0.503 $\pm$ 0.055                & \textbf{0.058} $\pm$ \textbf{0.002}               & 0.529 $\pm$ 0.041 & \textbf{0.057} $\pm$ \textbf{0.003}       \\
    & Transformer AE               & 0.206 $\pm$ 0.001       & 0.058 $\pm$ 0.001                & \textbf{0.218} $\pm$ \textbf{0.002}                & 0.028 $\pm$ 0.002                & \textbf{0.496} $\pm$ \textbf{0.050}       & 0.061 $\pm$ 0.002               & 0.545 $\pm$ 0.037                 & 0.069 $\pm$ 0.013                 \\
    & Conv. Trans. AE & 0.208 $\pm$ 0.002                & 0.057 $\pm$ 0.001                & \textbf{0.218} $\pm$ \textbf{0.001}       & 0.057 $\pm$ 0.001                & 0.521 $\pm$ 0.059                & 0.068 $\pm$ 0.013              & 0.533 $\pm$ 0.035                & 0.062 $\pm$ 0.011                \\
    \cmidrule(r){1-10}
    \parbox[t]{1mm}{\multirow{4}{*}{\rotatebox[origin=c]{90}{\emph{MSE+SAD}}}}
    & Dense AE                     & 0.242 $\pm$ 0.008                  & \underline{\textbf{0.003}} $\pm$ \underline{\textbf{0.001}}                & 0.265 $\pm$ 0.040       & \textbf{0.003} $\pm$ \textbf{0.002}       & 0.498 $\pm$ 0.038 & 0.070 $\pm$ 0.006     & \textbf{0.509} $\pm$ \textbf{0.045} & 0.069 $\pm$ 0.008  \\
    & Convolutional AE             & 0.264 $\pm$ 0.068       & \textbf{0.004} $\pm$ \textbf{0.004}       & 0.240 $\pm$ 0.007 & \underline{\textbf{0.002}} $\pm$ \underline{\textbf{0.000}} & 0.568 $\pm$ 0.057                 & 0.083 $\pm$ 0.017              & 0.593 $\pm$ 0.052       & 0.080 $\pm$ 0.013       \\
    & Transformer AE               & 0.234 $\pm$ 0.006                  & \textbf{0.004} $\pm$ \textbf{0.001}                & 0.240 $\pm$ 0.009                & 0.004 $\pm$ 0.001                & 0.606 $\pm$ 0.034                & 0.101 $\pm$ 0.026              & 0.632 $\pm$ 0.014                & 0.097 $\pm$ 0.028                \\
    & Conv. Trans. AE & 0.321 $\pm$ 0.121                & 0.008 $\pm$ 0.005                & 0.333 $\pm$ 0.123                & 0.007 $\pm$ 0.006                & 0.570 $\pm$ 0.068       & 0.113 $\pm$ 0.012              & 0.665 $\pm$ 0.179                & 0.164 $\pm$ 0.037                \\
    \bottomrule
    \end{tabular}
    \caption{
    \textbf{Full unmixing results on Raman spectroscopy data from sugar solutions.} 
    {\normalfont Each value represents the average result of $5$ replicates. Confidence intervals are calculated as one standard deviation around the sample mean. Values rounded to $3$ decimal places. Best and second best results (mean value) are given in \underline{\textbf{underlined bold}} and \textbf{bold}, respectively.
    }
    }
    \label{tab:sugar_metrics_full}
\end{table*}

\begin{table*}[!h]
    \centering
    \tiny
    \begin{tabular}{llcccccccc}
    \toprule
    & & \multicolumn{4}{c}{\emph{High SNR}}& \multicolumn{4}{c}{\emph{Low SNR}}
    \\\cmidrule(r){3-6}\cmidrule(r){7-10}   
    & & \multicolumn{2}{c}{\emph{with endmembers}} & \multicolumn{2}{c}{\emph{without endmembers}} & \multicolumn{2}{c}{\emph{with endmembers}} & \multicolumn{2}{c}{\emph{without endmembers}} 
    \\\cmidrule(r){3-4}\cmidrule(r){5-6}\cmidrule(r){7-8}\cmidrule(r){9-10}   
    & & Endmembers & Abundances & Endmembers & Abundances & Endmembers & Abundances & Endmembers & Abundances\\
    & Method & (PCC) $\downarrow$ & (MSE) $\downarrow$ & (PCC) $\downarrow$ & (MSE) $\downarrow$ & (PCC) $\downarrow$ & (MSE) $\downarrow$ & (PCC) $\downarrow$ & (MSE) $\downarrow$\\
    \toprule
    & PCA                          & 0.547 $\pm$ 0.000                 & 0.073 $\pm$ 0.000                 & 0.541 $\pm$ 0.000                 & 0.072 $\pm$ 0.000                 & 0.641 $\pm$ 0.000                 & 0.075 $\pm$ 0.000               & 0.691 $\pm$ 0.000                  & 0.073 $\pm$ 0.000                 \\
    & N-FINDR+FCLS               & 0.039 $\pm$ 0.000   & 0.005 $\pm$ 0.000         & 0.203 $\pm$ 0.055                & 0.037 $\pm$ 0.029 & 0.520 $\pm$ 0.000                   & \underline{\textbf{0.026}} $\pm$ \underline{\textbf{0.000}} & 0.641 $\pm$ 0.000                 & 0.154 $\pm$ 0.000                 \\
    & VCA+FCLS                   & 0.099 $\pm$ 0.047                & \textbf{0.004} $\pm$ \textbf{0.000}   & 0.221 $\pm$ 0.056                & 0.042 $\pm$ 0.029       & 0.325 $\pm$ 0.019                & 0.077 $\pm$ 0.035              & 0.482 $\pm$ 0.005                & 0.108 $\pm$ 0.006                \\
    \cmidrule(r){1-10}
    \parbox[t]{1mm}{\multirow{4}{*}{\rotatebox[origin=c]{90}{\emph{SAD}}}} & Dense AE                     & 0.029 $\pm$ 0.001                & 0.057 $\pm$ 0.001                & \textbf{0.031} $\pm$ \textbf{0.000}                 & 0.058 $\pm$ 0.001                & \underline{\textbf{0.164}} $\pm$ \underline{\textbf{0.042}} & \textbf{0.058} $\pm$ \textbf{0.005}     & \textbf{0.198} $\pm$ \textbf{0.046}       & \underline{\textbf{0.054}} $\pm$ \underline{\textbf{0.007}} \\
    & Convolutional AE             & \underline{\textbf{0.027}} $\pm$ \underline{\textbf{0.000}}       & 0.058 $\pm$ 0.001                & \underline{\textbf{0.030}} $\pm$ \underline{\textbf{0.002}} & 0.058 $\pm$ 0.001                & 0.207 $\pm$ 0.052                & \textbf{0.058} $\pm$ \textbf{0.002}               & 0.228 $\pm$ 0.030 & \textbf{0.057} $\pm$ \textbf{0.003}       \\
    & Transformer AE               & \textbf{0.028} $\pm$ \textbf{0.000}       & 0.058 $\pm$ 0.001                & \textbf{0.031} $\pm$ \textbf{0.000}                & 0.028 $\pm$ 0.002                & 0.200 $\pm$ 0.048       & 0.061 $\pm$ 0.002               & 0.243 $\pm$ 0.032                 & 0.069 $\pm$ 0.013                 \\
    & Conv. Trans. AE & \textbf{0.028} $\pm$ \textbf{0.000}                & 0.057 $\pm$ 0.001                & \textbf{0.031} $\pm$ \textbf{0.000}       & 0.057 $\pm$ 0.001                & 0.226 $\pm$ 0.059                & 0.068 $\pm$ 0.013              & 0.236 $\pm$ 0.034                & 0.062 $\pm$ 0.011                \\
    \cmidrule(r){1-10}
    \cmidrule(r){1-10}
    \parbox[t]{1mm}{\multirow{4}{*}{\rotatebox[origin=c]{90}{\emph{MSE+SAD}}}}
    & Dense AE                     & 0.037 $\pm$ 0.002                  & \underline{\textbf{0.003}} $\pm$ \underline{\textbf{0.001}}                & 0.047 $\pm$ 0.018       & \textbf{0.003} $\pm$ \textbf{0.002}       & \textbf{0.174} $\pm$ \textbf{0.032} & 0.070 $\pm$ 0.006     & \underline{\textbf{0.185}} $\pm$ \underline{\textbf{0.042}} & 0.069 $\pm$ 0.008  \\
    & Convolutional AE             & 0.056 $\pm$ 0.044       & \textbf{0.004} $\pm$ \textbf{0.004}       & 0.037 $\pm$ 0.002 & \underline{\textbf{0.002}} $\pm$ \underline{\textbf{0.000}} & 0.243 $\pm$ 0.044                 & 0.083 $\pm$ 0.017              & 0.272 $\pm$ 0.040       & 0.080 $\pm$ 0.013       \\
    & Transformer AE               & 0.035 $\pm$ 0.002                  & \textbf{0.004} $\pm$ \textbf{0.001}                & 0.037 $\pm$ 0.002                & 0.004 $\pm$ 0.001                & 0.296 $\pm$ 0.041                & 0.101 $\pm$ 0.026              & 0.311 $\pm$ 0.020                & 0.097 $\pm$ 0.028                \\
    & Conv. Trans. AE & 0.104 $\pm$ 0.096                & 0.008 $\pm$ 0.005                & 0.108 $\pm$ 0.099                & 0.007 $\pm$ 0.006                & 0.250 $\pm$ 0.072       & 0.113 $\pm$ 0.012              & 0.348 $\pm$ 0.176                & 0.164 $\pm$ 0.037                \\
    \bottomrule
    \end{tabular}
    \caption[]{
    \textbf{Full unmixing results on Raman spectroscopy data from sugar solutions with respect to an alternative endmember similarity metric PCC based on the Pearson correlation coefficient.} 
    {\normalfont Each value represents the average result of $5$ replicates. Confidence intervals are calculated as one standard deviation around the sample mean. Values rounded to $3$ decimal places. Best and second best results (mean value) are given in \underline{\textbf{underlined bold}} and \textbf{bold}, respectively. The PCC between two endmember signatures $x$ and $y$ is defined as $\displaystyle\text{PCC}(x, y) = 1 - \frac{\sum{}{}(x_i-\mean{x})\sum{}{}(y_i-\mean{y})}{\sqrt{\sum{}{}(x_i-\mean{x})^2\sum{}{}(y_i-\mean{y})^2}}$.} 
    }
    \label{tab:sugar_metrics_ncc}
\end{table*}

\end{document}